\documentclass{article}

    \usepackage[final]{neurips_2025}

\usepackage[utf8]{inputenc} 
\usepackage[T1]{fontenc}    
\usepackage{hyperref}       
\usepackage{url}            
\usepackage{booktabs}       
\usepackage{amsfonts}       
\usepackage{nicefrac}       
\usepackage{microtype}      
\usepackage{xcolor}         
\usepackage{float}          
\usepackage{graphicx} 
\usepackage{amsmath}
\usepackage{subcaption}
\usepackage[most]{tcolorbox}

\title{AutoHood3D: A Multi‑Modal Benchmark for Automotive Hood Design and\\ Fluid–Structure Interaction}

\author{%
  Vansh Sharma\thanks{Corresponding Author: vanshs@umich.edu} \\
  University of Michigan \\
  Ann Arbor, MI, USA \\
  \And
  Harish Jai Ganesh \\
  University of Michigan \\
  Ann Arbor, MI, USA \\
  \And
  Maryam Akram \\
  Ford Research and Innovation Center \\
  Dearborn, MI, USA \\
  \And
  Wanjiao Liu \\
  Ford Research and Innovation Center \\
  Dearborn, MI, USA \\
  \And
  Venkat Raman \\
  University of Michigan \\
  Ann Arbor, MI, USA \\
}

\begin{document}
\maketitle

\begin{abstract}
    This study presents a new high-fidelity multi-modal dataset containing 16000+ geometric variants of automotive hoods useful for machine learning (ML) applications such as engineering component design and process optimization, and multiphysics system surrogates. The dataset is centered on a practical multiphysics problem—hood deformation from fluid entrapment and inertial loading during rotary‑dip painting. Each hood is numerically modeled with a coupled Large-Eddy Simulation (LES)-Finite Element Analysis (FEA), using 1.2M cells in total to ensure spatial and temporal accuracy. The dataset provides time-resolved physical fields, along with STL meshes and structured natural language prompts for text-to-geometry synthesis. Existing datasets are either confined to 2D cases, exhibit limited geometric variations, or lack the multi‑modal annotations and data structures—shortcomings we address with AutoHood3D. We validate our numerical methodology, establish quantitative baselines across five neural architectures, and demonstrate systematic surrogate errors in displacement and force predictions. These findings motivate the design of novel approaches and multiphysics loss functions that enforce fluid–solid coupling during model training. By providing fully reproducible workflows, AutoHood3D enables physics‑aware ML development, accelerates generative‑design iteration, and facilitates the creation of new FSI benchmarks. Dataset and code URLs in \ref{sec:links}.
\end{abstract}

\section{Introduction}
Classical numerical solvers \cite{jasak2009openfoam, sharma2024amrex} provide accurate predictions for practical systems such as propulsion systems \cite{houtman-2023, seraj-2023, abisleiman2025structure} and turbine performance \cite{nassini-2021_turbine, chmielewski-2020_turbine, zhang2025three}; however, they are computationally expensive and often intractable for iterative design processes \cite{yagoubi2024neurips2024ml4cfdcompetition}. The latest advances in computational fluid dynamics (CFD) now incorporate ML techniques to develop turbulence closure models \cite{boral2023neural, van2024energy}, modify wall-modeled LES \cite{maulik2019sub}, and for real-time flow control strategies \cite{wei2024diffphycon, morton2018deep}. Traditional ML approaches have improved both predictive accuracy \cite{Petros2022multiscale, sharma2025machine} and time-to-solution across a variety of CFD applications \cite{kochkov2021machine, sharma2025accelerating}. Despite these advances, iterative design workflows rely predominantly on design-of-experiments methodologies integrated with CFD solvers to drive large‐scale optimization studies. 

In recent years, several comprehensive aerodynamic datasets \cite{ling2022blastnet, ohana2024thePDEwell} have been published, covering everything from 2D airfoil profiles \cite{bonnet2022airfrans} to full 3D vehicle exteriors \cite{elrefaie2024drivaernet++} and complete car geometries \cite{ashton2024windsor}. These large-scale collections serve to bridge the gap between data-driven surrogate models and high-fidelity CFD solvers. However, most publicly available repositories are RANS based, confined to 2D configurations, or lack both a sufficient sample size and a transparent generation workflow to support in‐depth design exploration. 
Moreover, real-world applications frequently involve additional physics, such as multiphase flows or coupled fluid-structure interactions, that these datasets do not capture. To date, no resource offers a reproducible, end-to-end pipeline alongside a large-scale, multiphysics 3D dataset.

\begin{figure}[!htb]
  \centering
  \includegraphics[width=0.90\linewidth]{ 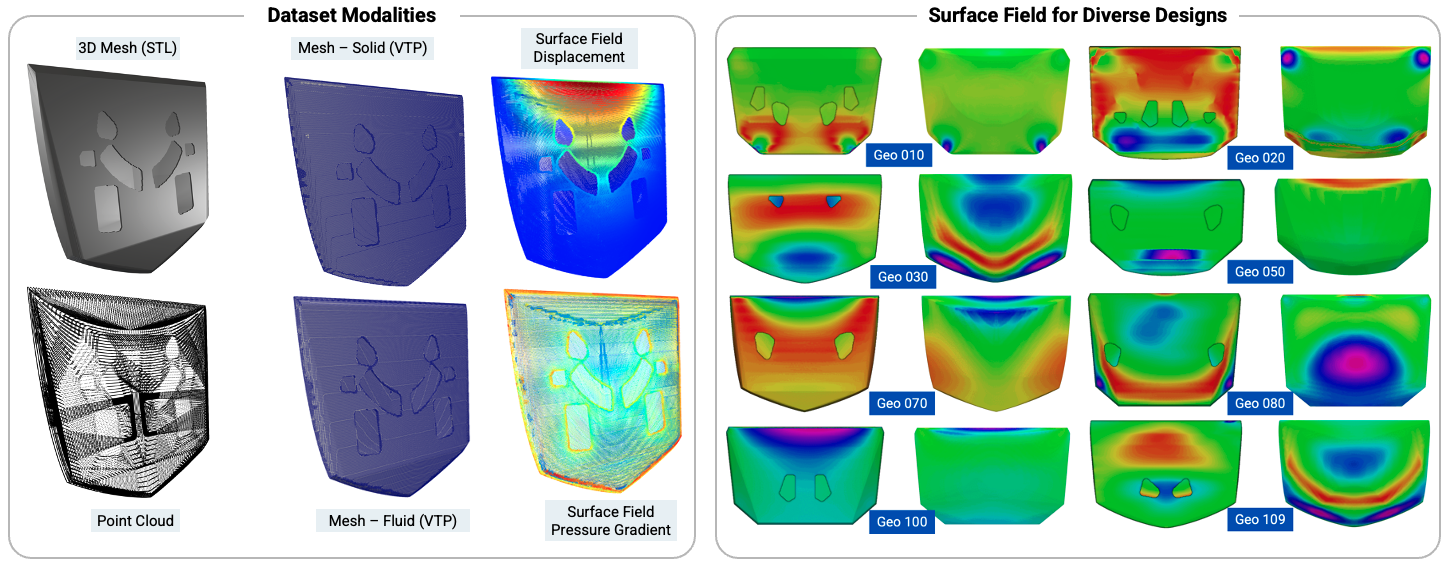}
  \caption{Dataset modalities (left): base 3D STL hood shell, surface point‐cloud sampling, raw fluid and solid meshes, and sample surface fields (normalized displacement and pressure‐gradient) extracted on the hood surface. Example variants (right): selected hood geometries with interpolated deformation fields mapped onto the STL, demonstrating how different cutout topologies yield distinct deflection patterns on both the front and rear faces.}
  \label{fig:dataset_mods}
\end{figure}

We present the AutoHood3D dataset (shown in Figure \ref{fig:dataset_mods}) to overcome the aforementioned gaps. In particular, the dataset poses a unique challenge for current data-driven methods due to its multiphysics modality. The multiphysics challenge at the core of this study is the deformation of automotive hoods during the rotary dip paint process \cite{kim2015predictionHood, kim2014microhood}. The automotive hood consists of two principal elements: an exterior panel designed for aerodynamic flow management and aesthetic styling, and an interior reinforcement, typically a hollow honeycomb lattice that accommodates fluid lines and provides thermal insulation while resisting static and dynamic deflection. Aerodynamic efficiency remains a key design factor, but pedestrian safety regulations \cite{shojaeefard-2013PedSaftey, yang-2021PedSafety} and utility requirements impose orthogonal constraints on the overall hood design. However, during the rotary dip paint process, rapid rotation of the carrier arm entrains residual coating fluid within the engineered cutouts and recesses, subjecting the shell to transient body forces that, combined with inertial loading, produce localized deflections \cite{kim2015predictionHood}. Because these manufacturing-induced loads are rarely incorporated into early-stage design analyses, unexpected deformations occur in thin-walled, geometrically complex regions. This behavior is analogous to industrial occurrences, such as, distortion of composite panels from immersion coatings, and finds a conceptual parallel in adaptive warping wing designs \cite{thill-2008Warping}.

The dataset comprises over 16,000 unique 3D hood geometries with time-resolved FSI solution snapshots, provided in multiple modalities (see Figure \ref{fig:dataset_mods}). All geometry generation scripts, FSI simulation workflows, and analysis tools are released under an open-source license and readily available to the developer community. We expect these modular workflows to accelerate the creation of new large-scale 3D FSI benchmarks, enabling temporal autoregressive modeling and physics-informed \cite{toscano2025pinns} PDE learning applications.
In an overarching way, our contributions can be summarized as follows:
\begin{itemize}
  \item Introduce the first open-source dataset of more than 16,000 3D automotive hood geometries with engineered cutouts, providing extensive design diversity for fluid dynamics and structural studies.
  
  \item Provide a fully reproducible, end-to-end workflow, from STL generation to FSI co-simulation, designed for scalable synthesis of design variants of new components, and extensible to diverse engineering contexts. 
  
  \item Deliver a truly multimodal benchmark, comprising high-fidelity STL meshes, raw CFD solutions with time-resolved flow and surface fields, FEA displacement and stress outputs, surface point clouds, and tokenized natural language annotations, accompanied by performance benchmarks for five neural architectures.
\end{itemize}

\section{Related Work}
In scientific‐ML (SciML) domain, the integration of AI into CFD to accelerate high‐fidelity simulations has become a prominent research thrust. Large‐scale repositories, such as PDEBench \cite{takamoto2022pdebench}, PINNacle \cite{zhongkai2024pinnaclePDE}, Airfrans \cite{bonnet2022airfrans}, and BubbleML \cite{hassan2023bubbleml}, provide comprehensive datasets for PDE learning, airfoil optimization, and multiphase flow modeling. Beyond application‐specific datasets, cross‐domain databases such as BLASTNet \cite{ling2022blastnet} and ThePDEWell \cite{ohana2024thePDEwell} aggregate varied PDE system results and spatio‐temporal solution fields, enabling systematic benchmarking and transferability studies across scientific disciplines. In the industrial domain, large‐scale datasets such as DrivAerNet++ \cite{elrefaie2024drivaernet++} and WindsorML \cite{ashton2024windsor} have emerged, presenting realistic scenarios that enable engineers to predict surface pressure distributions, volumetric flow characteristics, and the resulting forces and moments on novel geometries under varied operating conditions. 
Despite their individual strengths, existing datasets exhibit notable gaps: some provide extensive geometric and solution data but lack generative workflows for creating new variants \cite{zhongkai2024pinnaclePDE, ohana2024thePDEwell, ling2022blastnet}; others deliver end-to-end pipelines, yet remain confined to 2D cases with only limited extensions to three dimensions \cite{bonnet2022airfrans, hassan2023bubbleml}; and a few support full 3D flow simulations but offer limited design diversity due to computational costs \cite{ashton2024windsor}. Consequently, no public dataset unifies broad design variations, reproducible generation processes, and a comprehensive 3D FSI benchmark with multiple modalities such as point cloud and STL. 

In parallel, recent work \cite{thengane2025foundationalPoint} has shown that point clouds serve as a versatile representation for physical world and scene reconstruction \cite{liu2024uni3d} across applications such as virtual reality, object recognition, autonomous navigation \cite{mao20233dPlanning}, and robotic manipulation. For example, PointLLM \cite{xu2024pointllm} and related models \cite{tang2024minigpt_pointcloud} use joint text–point‑cloud embeddings \cite{guo2023pointJointEmbs} to generate 3D geometries from natural language prompts \cite{lahoud20223dPointLLMs_survey}. Despite these advances \cite{deitke2023objaversexluniverse10m3d, lin2025objaversecurated3dobject}, the development of text‐grounded CAD point cloud models is hindered by the absence of large‑scale, diverse 3D engineering datasets with detailed natural language annotations. Furthermore, point cloud data are inherently sparse and typically lack semantic context, limiting their utility for comprehensive scene understanding \cite{wang2024unibev, thengane2025foundationalPoint}. To bridge these domains, we augment our FSI dataset with paired natural language annotations and point cloud outputs, creating a text‑to‑geometry corpus for supervised fine‑tuning of LLMs in CAD‑driven generative‑AI workflows.

\section{Dataset Workflow}
The dataset generation workflow is organized into three broad phases. In the first phase we process the inner–hood designs from \cite{carHooks10k} to reconstruct complete hood shell and segment cutouts as ordered point clouds. In the second phase focused on design‐variation synthesis, we generate new hood variants by embedding the previously extracted cutouts chosen under user‐defined spacing and plane‐separation constraints. Finally, in the third phase, consisting of the FSI simulation pipeline, we automate mesh generation and solver initialization to perform fluid–structure analyses at scale. Detailed methodologies are as follows.

\subsection{Base Geometry Generation}
\begin{figure}[h]
  \centering
  \includegraphics[width=1.0\linewidth]{ 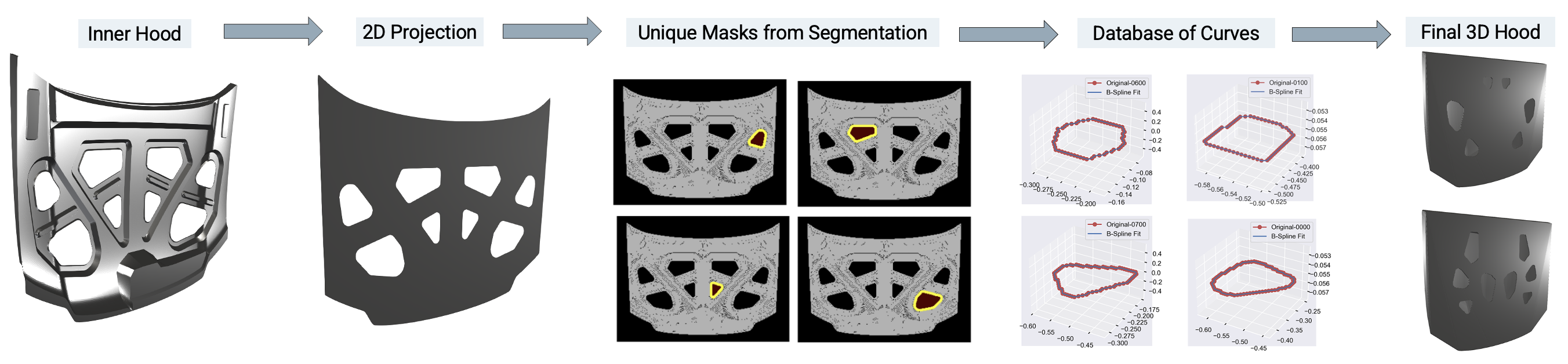}
  \caption{Workflow for generating multiple hood geometries. Starting from the base inner‐hood CAD (left), the surface is projected onto a 2D plane and segmented into individual cutout masks. Each mask boundary is extracted as an ordered point-cloud curve. These curves are then re-embedded into the full 3D hood shell to produce the final geometries with engineered openings (right).}
  \label{fig:cuts}
\end{figure}

We adopt 100 inner‐only hood geometries from \cite{carHooks10k} as our base set. To obtain a complete dual‐shell model, each inner‐hood design is processed with a convex‐hull algorithm \cite{pymadcad} to produce an outer envelope surface. The resulting hull is then translated and rotated so that the inner shell is centered at the global origin and aligned approximately perpendicular to the incoming flow direction. Finally, a uniform thickness of 10 mm is extruded from the hull to form the full outer and inner hood shell. To generate a diverse set of engineering cutouts on these base shells, we draw on a collection of approximately 250 inner‐hood geometries from \cite{carHooks10k}, each of which contains about four to eight cutouts. Each feature of the cutout is parameterized by smooth Bézier curves, thus avoiding sharp corners that can induce stress concentrations. To isolate these cutouts curves, we project the 3D inner hood surface onto a reference plane and employ the SAM-2 segmentation model \cite{ravi2024sam2} to delineate individual cutout zones (see Fig. \ref{fig:cuts}). From each segmented region, the curve boundary is extracted to produce an ordered point cloud, which is then stored in a design database.

The design database comprises approximately 1,750 unique cutout profiles, each characterized by its perimeter and enclosed area. We apply the K-Means \cite{kmeans} clustering algorithm in the two‐dimensional feature space (perimeter, area), which yields nine distinct groups (see Fig. \ref{fig:clusters}). When generating new hood variants, we sample between one and four curves from the unique clusters, enforcing bilateral symmetry about the vehicle (or hood) centerline. Two user‐defined parameters govern the layout: (1) the minimum distance between adjacent curves on a given cutting plane, and (2) the axial separation between mirrored cutout planes. For each design iteration, we uniformly sample these parameters from prescribed ranges to ensure both engineering relevance and sufficient diversity in the resulting hood geometries. Additional information provided in Appendices \ref{appendix:hoodVariations} and \ref{appendix:clustering}

\subsection{Solver Setup}
\begin{figure} [!htb]
  \centering
  \includegraphics[width=0.85\linewidth]{ 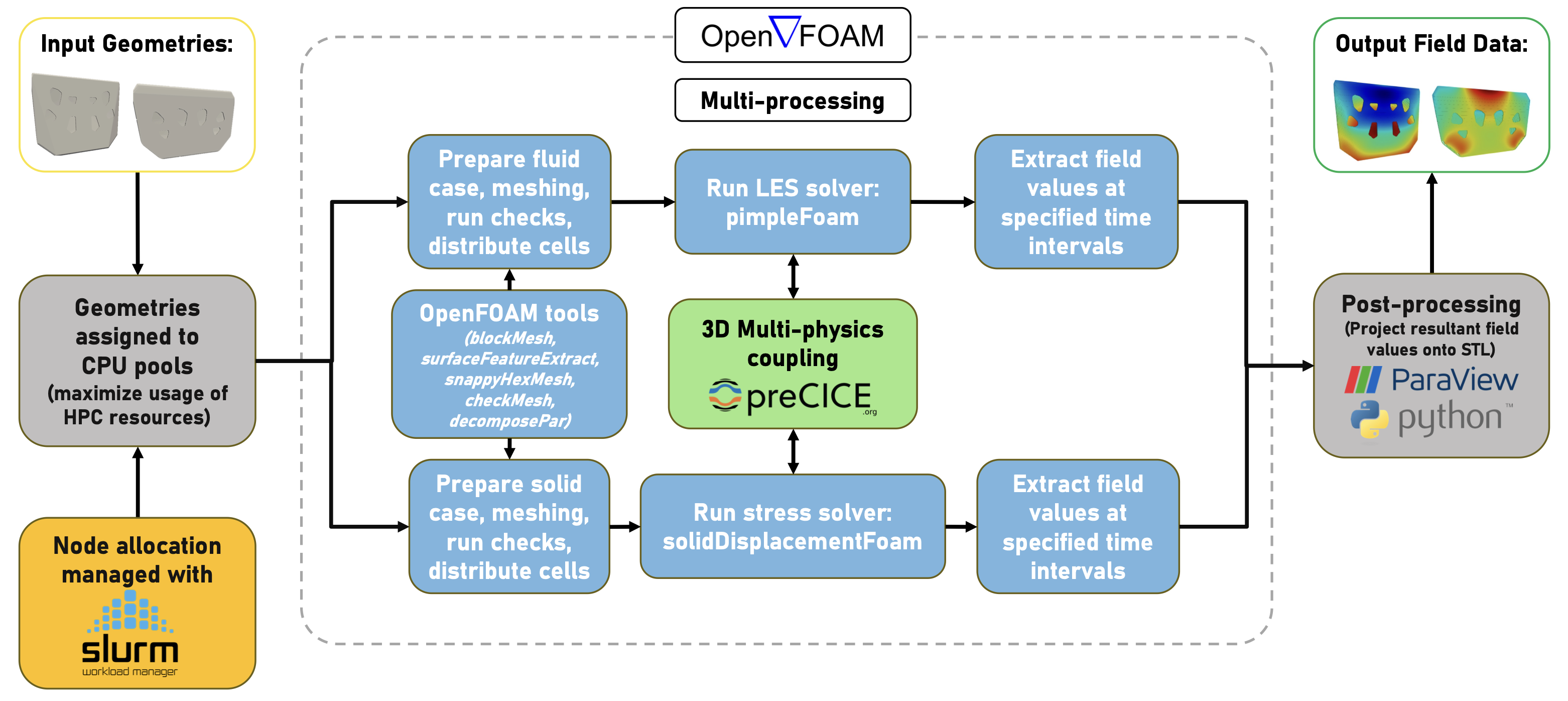}
  \caption{Co-simulation workflow.}
  \label{fig:co_sim_over}
\end{figure}
All simulations were carried out using the open‐source OpenFOAM v2312 framework \cite{OpenFOAMESI}. For the fluid phase, we developed UM\_pimpleFoam, a CPU‐based LES solver that extends OpenFOAM’s transient, incompressible pimpleFoam solver with the Spalart–Allmaras Delayed Detached‐Eddy‐Simulation (DDES) turbulence closure \cite{spalart2006new}. The primary modification in UM\_pimpleFoam is the inclusion of an additional transient forcing term in the momentum‐conservation equation, enabling explicit time‐dependent body‐force effects within the Navier–Stokes system. For the structural phase, we used UM\_solidDisplacementFoam, built on top of OpenFOAM’s solidDisplacementFoam utility. This solver assumes a small-strain linear-elastic constitutive model, Hooke's law, to compute the hood stress and displacement fields.  
Since the predicted deformations are minor and exert negligible spatio‐temporal perturbations on the flow field, we adopt a one-way coupling approach: the fluid solver drives the structural response without reverse feedback. Multiple studies have shown that one-way coupling can capture the correct system response \cite{shinde-2023, kimmel-2024, benra-2011FSI, hagmeyer-2022FSI}. Critically, the fluid and solid solvers execute concurrently, exchanging pressure field at each coupling interval (Fig.~\ref{fig:co_sim_over}). Additional details on computational setup and solver test cases are provided in Appendix \ref{appendix:solvers}.

\begin{figure} [!htb]
  \centering
  \includegraphics[width=0.85\linewidth]{ 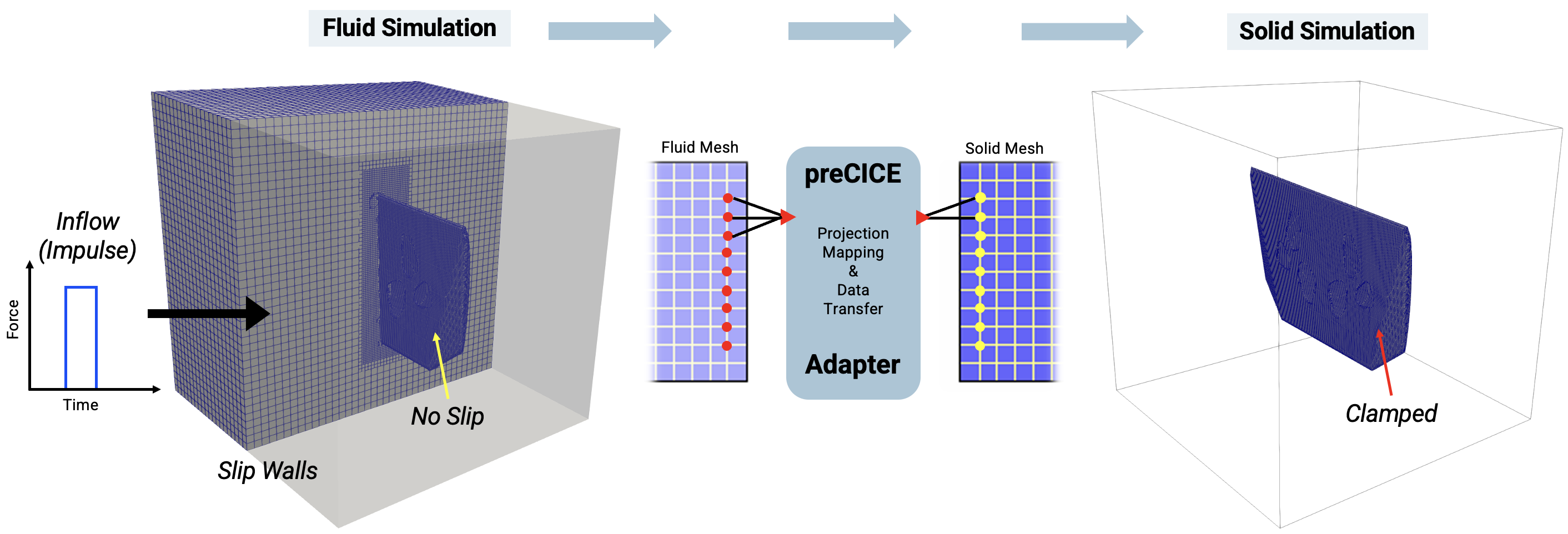}
  \caption{One-way coupled FSI co-simulation overview. An impulsive inflow acceleration is applied to the fluid domain, which uses free-slip exterior walls and a no-slip hood surface. The computed pressure field on the fluid mesh is transferred via preCICE’s nearest-projection adapter to the solid mesh, where the hood shell is fully clamped for structural deformation analysis.}
  \label{fig:co_sim_setup}
\end{figure}

\paragraph{Inter‐solver communication} preCICE library \cite{preCICEv2} provides the essential adapter to couple multiphysics simulations by mapping boundary value exchanges at each coupling interval \cite{OpenFOAMpreCICE}. We adopt a nearest‐projection scheme to map the fluid‐solver pressure field onto the structural-solver mesh (shown in Fig. \ref{fig:co_sim_setup}) and to transfer deformation if bidirectional coupling were to be employed in future. Because both meshes are generated from an identical STL source, the geometric discrepancy is negligible, ensuring that the nearest‐projection operator introduces minimal interpolation error.

\paragraph{Computational Mesh}
Mesh generation was performed with OpenFOAM’s SnappyHexMesh utility, incorporating targeted surface boundary refinements to capture near‐wall flow features \cite{heft-2012, ashton2024windsor}. The final LES mesh comprises approximately 750,000 polyhedral cells ($\Delta$ = 0.003125 m) in total, of which 170,000 to 180,000 cells are concentrated in the boundary layer region surrounding the surface of the hood to ensure accurate capture of pressure distribution phenomena. For structural analysis, a surface mesh of approximately 400,000 cells ($\Delta$ = 0.00220 m) was extracted from the same STL geometry, providing sufficient spatial resolution to resolve the dominant deflection mode \cite{garhuom2020FEA}.

\paragraph{Boundary Conditions}
The computational domain is tightly wrapped around the hood to concentrate the resolution on the fluid loads acting on its surface. At the downstream boundary, a zero-gradient (extrapolation) outflow condition is imposed to suppress spurious reflections back into the domain.
The hood geometry is placed normal to the incoming flow, with its engineered cutouts facing upstream. We apply a time-dependent acceleration impulse at the inlet, that is, a uniform force history imposed simultaneously on all fluid cells, while the flow is initialized in a quiescent state. For all simulations, a constant acceleration pulse of magnitude $\alpha$ = 2.7 m/s$^2$ is applied over a fixed duration of $t$ = 0.07 s at 30\textdegree ~to the flow direction (sampling time 0.01s, total of 8 solutions). The structural domain is defined as the complement of the fluid mesh (with only the hood shell), and is fully clamped at its mounting interfaces with material properties set as: density ($\rho_s$) = 2700 kg/m$^3$, Poison ratio ($\nu$) = 0.33 and Young’s modulus ($E$) = 68.9e9 N/m. Figure \ref{fig:co_sim_setup} details the boundary conditions and domain configuration used in the co‐simulation. In the fluid domain, all faces have slip or ``inviscid'' conditions and has no fluid boundary layer growth; the hood is treated as a no-slip wall to capture viscous effects at the solid–fluid interface.

\paragraph{HPC Setup} Simulations were executed on our internal HPC cluster, comprising 16 compute nodes with 8 NVIDIA H100 SXM 80GB GPUs each and a total of 1536 CPU cores (Intel Xeon Platinum 8468 2.1G). Each simulation case was parallelized over 12 cores with 72 GB of memory requirement. The full simulation dataset occupies approximately 1.2 TB of storage and consists of 37,000 output files. In aggregate, the co-simulation campaign consumed roughly 5000 CPU-core hours.

\subsection{Dataset for Generative AI}
\begin{figure} [!htb]
  \centering
  \includegraphics[width=0.8\linewidth]{ 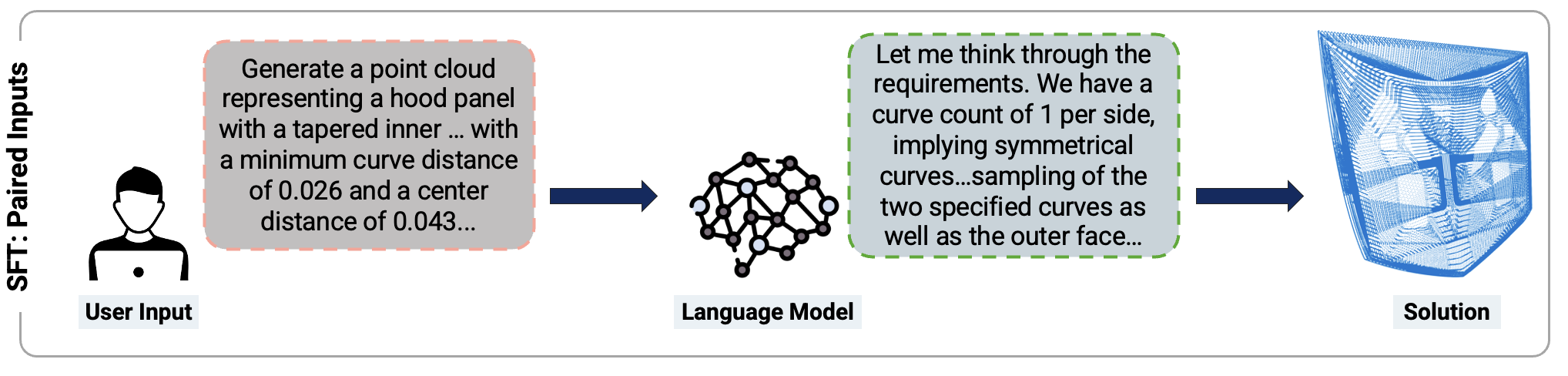}
  \caption{Supervised fine‑tuning (SFT) example data point: an LLM is trained to map structured text prompts, specifying cutout parameters such as curve count and geometric spacing, to the corresponding 3D point cloud hood geometry.}
  \label{fig:llm}
\end{figure}
We constructed a text-based paired corpus of 2587 hood geometries by leveraging the Gemma3 model (gemma3-27b-it)\cite{team2025gemma} in its vision‐language configuration. For each sample, we authored a concise text prompt encoding the engineered cutout parameters: minimum intercurve spacing, center plane offset, and total curve count along with a snapshot of the geometry and passed it through Gemma3’s multimodal encoder. The model then generated the corresponding 3D point cloud description of the hood shell and a chain-of-thought procedure, which we captured as the supervised output (see Fig. \ref{fig:llm}). This dataset supports training and supervised fine‑tuning (SFT) of LLMs to map natural language design specifications onto 3D point cloud representations, thus establishing a foundation for generative AI CAD workflows ranging from zero shot shape synthesis and parametric editing to language‑driven mesh refinement and automated design variation generation. Additional details in SI.

\subsection{Dataset Structure}
The dataset encompasses automotive hood designs across multiple modalities, ensuring diverse representations and enabling seamless integration into a variety of downstream ML and generative AI applications.
\begin{itemize}
  \item 3D Base Geometries: High-fidelity STL meshes of complete hood shells, suitable for CAD workflows, CFD mesh generation, design optimization, and generative modeling.
  \item CFD Results: Time-resolved surface-interpolated velocity and pressure fields at eight temporal snapshots, ideal for data-driven fluid dynamics studies and autoregressive sequence models.
  \item FEA Results: Displacement and stress fields on the hood under CFD-derived loads, allowing structural performance analysis and surrogate model training.
  \item Point Clouds: Surface point cloud representations of the hood with variable point densities, for use in reconstruction algorithms.
  \item Engineered STL Variations: A diverse set of STL hood shells with integrated cutouts, curated for generative-design, topology-optimization, and virtual-reality applications.
  \item Text Descriptions: Tokenized natural language annotations for each STL file, facilitating transformer-based mapping between textual metadata and geometric models.
\end{itemize}

The dataset is organized into two levels, a sample subset of 4500 (approximately) and an extended sample superset of 12000+ hoods, structured in a curriculum learning sequence \cite{bengio2009curriculum}. Beyond these datasets, an additional set consisting of 10000+ hood geometries with randomized cutout configurations is included. Within each tier, samples progress from designs with the fewest cutout curves to those with the most, enabling staged training from simpler to more complex geometric variations. Additional details in Appendix \ref{appendix:dataset}.

\paragraph{Dataset Access} The dataset is released under CC BY-NC-ND 4.0 and is freely available without requiring an account (see \ref{sec:links}); full download instructions and metadata (Croissant format) appear in SI and GitHub. Due to the large size of the data, pre-split subsets for training, validation, in-distribution (id) testing, and out-of-distribution (ood) testing are also provided. 

\section{ML Benchmarking}
Accurate prediction of fluid‐induced deformations is essential in the automotive hood design process. Because rapid evaluation of numerous design variants, each subject to multiple performance constraints, is required, low-latency surrogate models are indispensable. In this section, building on previous benchmarks \cite{elrefaie2024drivaernet++, bonnet2022airfrans, ashton2024windsor}, we present ML approaches for predicting the normalized deformation response across different hood geometries.
Using PyTorch \cite{pytorch}, we evaluate five architectures—MLP, PointNet \cite{qi2017pointnet}, GraphSAGE \cite{hamilton2017inductiveGSAGE}, Graph U-Net \cite{gao2019graphUNet}, and PointGNNConv \cite{shi2020pointGCNNConv}—each composed of an encoder, core model, and decoder. An MLP-based encoder first maps each hood vertex (or point-cloud sample) and its auxiliary attributes (e.g., surface pressure) into a latent feature vector. The core model then aggregates these embeddings, via point- or graph-based operations, to capture local and global geometric context. Finally, an MLP-based decoder regresses the normalized deformation at each node. For this study, we used the 4500 hood geometries subset, partitioned into training (75\%), validation (15\%), id-testing (10\%), and ood-testing (single isolated geometry) segments. Model inputs consist of point‑cloud coordinates, surface normals, and mesh‑connectivity data, while outputs comprise the velocity components $U$, pressure $p$, and displacement $D$. All models were trained for 750 \footnote{\label{noteGU}Graph U‑Net results are reported at 320 epochs due to its longer per‑epoch runtime. However, the training loss curve aligns closely with those of the leading models trained for 750 epochs.} epochs using the mean squared error (MSE) loss function. Comprehensive hyperparameter configurations and architectural details are provided in Appendix \ref{appendix:hyperParms}.
\paragraph{Benchmarking Metric} Model performance is evaluated using the Mean Squared Error (MSE), defined as the average of the squared differences between predicted and true deformations. Due to its quadratic error term, the MSE is particularly sensitive to outliers in prediction errors. 

\subsection{Benchmarking Results\label{sec:benchmarking}}
\subsubsection{In-Distribution Tests}
\begin{table}[htbp]
\centering
\footnotesize 
\begin{tabular}{lccccccc}
\toprule[1.25pt]
\textbf{Model}
  & \multicolumn{1}{c}{\begin{tabular}[c]{@{}c@{}}\textbf{$U_x$}\\\textbf{($\times10^{-2}$)}\end{tabular}}
  & \multicolumn{1}{c}{\begin{tabular}[c]{@{}c@{}}\textbf{$U_y$}\\\textbf{($\times10^{-2}$)}\end{tabular}}
  & \multicolumn{1}{c}{\begin{tabular}[c]{@{}c@{}}\textbf{$U_z$}\\\textbf{($\times10^{-2}$)}\end{tabular}}
  & \multicolumn{1}{c}{\begin{tabular}[c]{@{}c@{}}\textbf{$p$}\\\textbf{($\times10^{-2}$)}\end{tabular}}
  & \multicolumn{1}{c}{\begin{tabular}[c]{@{}c@{}}\textbf{$D_x$}\\\textbf{($\times10^{-2}$)}\end{tabular}}
  & \multicolumn{1}{c}{\begin{tabular}[c]{@{}c@{}}\textbf{$D_y$}\\\textbf{($\times10^{-2}$)}\end{tabular}}
  & \multicolumn{1}{c}{\begin{tabular}[c]{@{}c@{}}\textbf{$D_z$}\\\textbf{($\times10^{-2}$)}\end{tabular}} \\
\midrule[0.5pt]
MLP          &  \textbf{0.25} &  0.27 &  \textbf{0.44} &  \textbf{0.37} & 2.80 &  0.51 & 0.76 \\
PointNet     &  0.31 &  \textbf{0.26} &  0.48 &  0.55 & \textbf{0.29}  & \textbf{0.40} &  \textbf{0.43} \\
GraphSAGE    &  4.89  &  1.31  & 2.04  & 4.22  & 14.88  & 3.98  &  6.86  \\
Graph U‑Net \footref{noteGU}  &  1.91 &   0.86 &  1.18 & 3.09 &  2.89  & 0.92 & 0.97 \\
PointGNNConv &   4.50 &   1.42 &   2.69 &   7.70 &   11.71  &   8.92 &   15.41 \\ 
\midrule[1.25pt]
\end{tabular}
\caption{ID Test: Mean squared error on the different normalized fields for all the models.}
\label{tab:mse_iTest}
\end{table}

\begin{figure} [!htb]
  \centering
  \includegraphics[width=0.95\linewidth]{ 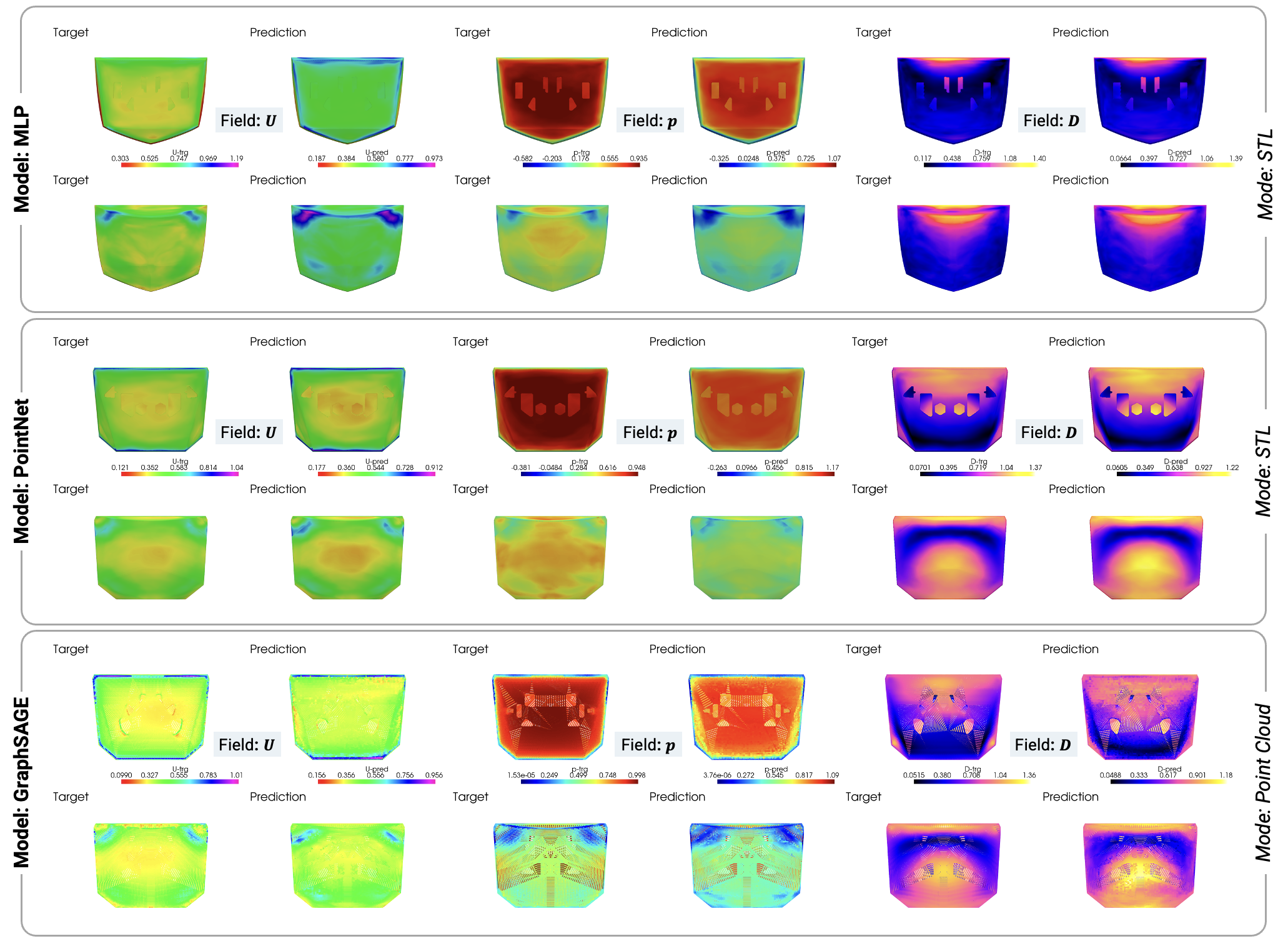}
  \caption{In‑distribution test predictions for the three lowest‑error models. Each block shows results for a specific mode—MLP (STL), PointNet (STL), and GraphSAGE (Point Cloud)—with target (left) and predicted (right) fields plotted for $\left\|U \right\|$ , $p$, and $\left\|D \right\|$  on the front (top row of each block) and back (bottom row of each block) surfaces of the hood geometry.}
  \label{fig:id_viz}
\end{figure}

Table \ref{tab:mse_iTest} and Fig. \ref{fig:id_viz} demonstrate that simple architectures, namely MLP and PointNet, outperform more complex graph-based methods on ID test data. As shown in the table, the MLP achieves the lowest MSE on $U_x$ and $U_z$, and on $p$, while PointNet shows the best overall balance of errors across all fields. The visualizations of the field map in Fig. \ref{fig:id_viz} confirm these quantitative findings. The MLP and PointNet reconstructions capture the primary patterns in $\left\|U\right\|$ and $p$ with minimal smoothing, and reproduce the high‐deflection zones around cutout edges in $\left\|D\right\|$. In contrast, graph-based models, such as GraphSAGE tend to underpredict peak values and introduce spurious oscillations in regions of sharp geometric change. Graph networks are based on a fixed neighborhood radius or neighbor count, which forces a trade-off between over-smoothing fine cutout details and depriving nodes of broader context. In particular, the dramatic degradation in $D_x$ performance is possibly due to displacement being a secondary field driven by the integrated pressure load, so any upstream errors in velocity or pressure predictions amplify downstream effects. See Appendix~\ref{appendix:mlp} for further analysis of MLP performance compared to graph models (homophily and ablation tests). 

\subsubsection{Out‐of‐Distribution Test}
\begin{figure} [!htb]
  \centering
  \includegraphics[width=0.95\linewidth]{ 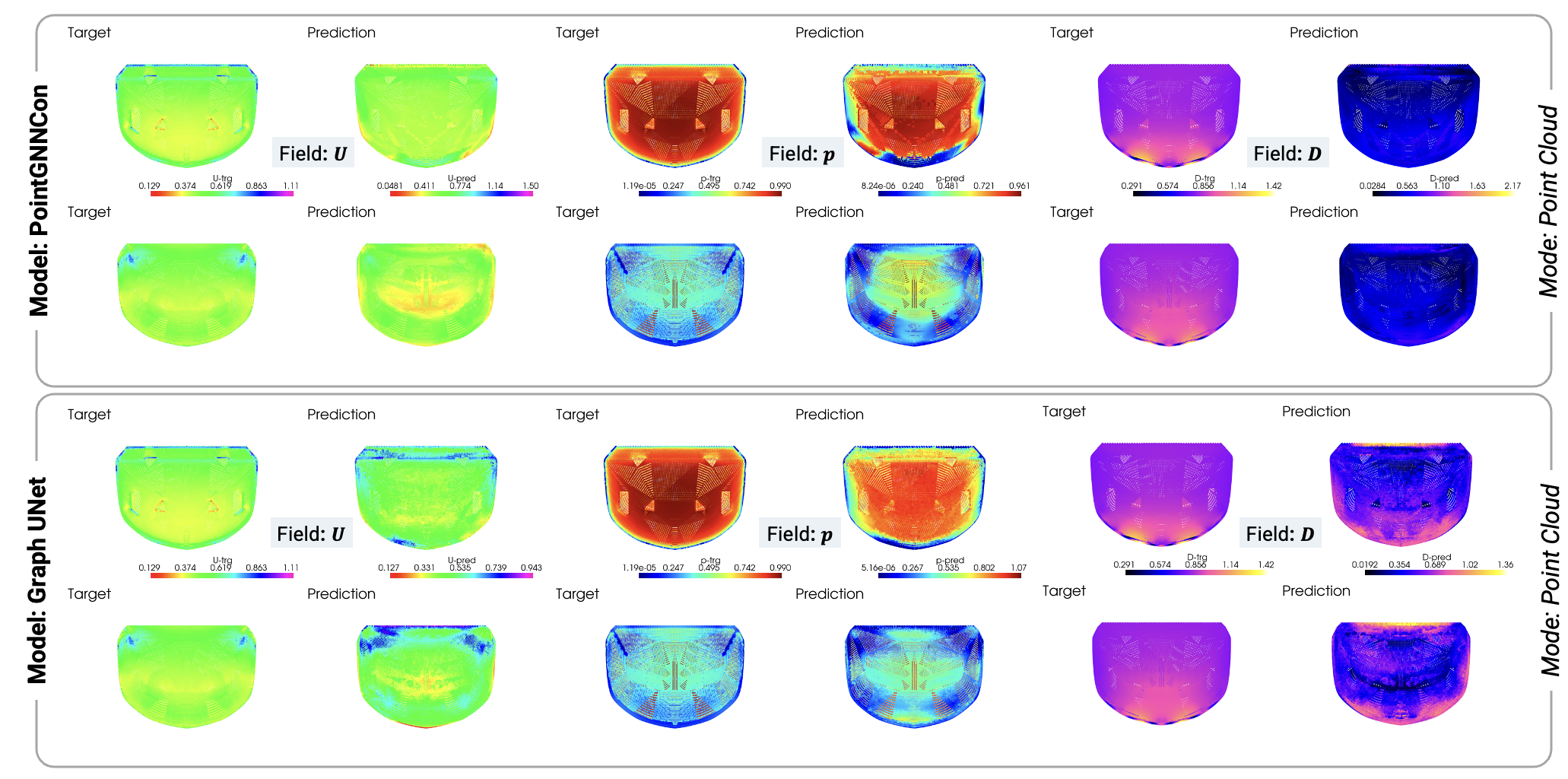}
  \caption{OOD test predictions for the top two graph‐based models using point‐cloud modality. For PointGNNConv (top block) and Graph U‑Net \footref{noteGU} (bottom block), target (left) and predicted (right) fields are shown for $\left\|U \right\|$, $p$, and $\left\|D \right\|$ on the front (upper row of each block) and back (lower row of each block) surfaces of a representative OOD hood geometry.}
  \label{fig:od_viz}
\end{figure}

\begin{table}[!htp]
\centering
\footnotesize 
\begin{tabular}{lccccccc}
\toprule[1.25pt]
\textbf{Model}
  & \multicolumn{1}{c}{\begin{tabular}[c]{@{}c@{}}\textbf{$U_x$}\\\textbf{($\times10^{-2}$)}\end{tabular}}
  & \multicolumn{1}{c}{\begin{tabular}[c]{@{}c@{}}\textbf{$U_y$}\\\textbf{($\times10^{-2}$)}\end{tabular}}
  & \multicolumn{1}{c}{\begin{tabular}[c]{@{}c@{}}\textbf{$U_z$}\\\textbf{($\times10^{-2}$)}\end{tabular}}
  & \multicolumn{1}{c}{\begin{tabular}[c]{@{}c@{}}\textbf{$p$}\\\textbf{($\times10^{-2}$)}\end{tabular}}
  & \multicolumn{1}{c}{\begin{tabular}[c]{@{}c@{}}\textbf{$D_x$}\\\textbf{($\times10^{-2}$)}\end{tabular}}
  & \multicolumn{1}{c}{\begin{tabular}[c]{@{}c@{}}\textbf{$D_y$}\\\textbf{($\times10^{-2}$)}\end{tabular}}
  & \multicolumn{1}{c}{\begin{tabular}[c]{@{}c@{}}\textbf{$D_z$}\\\textbf{($\times10^{-2}$)}\end{tabular}} \\
\midrule[0.5pt]
MLP          &  \textbf{1.49}  &  1.89  &  2.97  &  4.96  & 181.02 &  5.87  & 19.81  \\
PointNet     &  3.48  & 2.89  & 6.87  & 15.45  & \textbf{73.79} & 17.83  & 25.55  \\
GraphSAGE     &  3.24  &  1.81  &  2.65  &  6.99  & 84.34  & 6.07  & 21.71  \\
Graph U‑Net \footref{noteGU}  &  2.04 &  1.29 &  1.45 &  \textbf{4.05} &  126.62  &  7.69 &  16.13 \\
PointGNNConv &  2.84 &  \textbf{1.51} &  \textbf{1.20} &  5.79 &  130.08  &  \textbf{4.78} &  \textbf{13.67} \\ 
\midrule[1.25pt]
\end{tabular}
\caption{OOD Test: Mean squared error on the different normalized fields for all the models.}
\label{tab:mse_oTest}
\end{table}
The OOD test set reveals a different trend for model performance: graph‐based architectures outperform point‐ and MLP‐based models when faced with novel geometries (Table \ref{tab:mse_oTest}). While MLP achieves the lowest error on $U_x$ and PointNet on $D_x$, both exhibit substantial degradation in $p$ and $D$ predictions outside the training distribution. In contrast, Graph U-Net attains the best OOD MSE for pressure $4.05\times10^{-2}$ and maintains competitive errors across all velocity components, demonstrating a robust generalization of global flow characteristics. Similarly, PointGNNConv achieves the lowest error on transverse components $D_y$ and $D_z$, indicating its efficacy in capturing localized deformation induced by complex cutout topologies. The field‐map visualizations (Figure \ref{fig:od_viz}) confirm these quantitative results: Graph U-Net and PointGNNConv reproduce sharp features in pressure and the concentrated deflection zones around cutouts with high fidelity, whereas simpler models produce overly smooth or smeared predictions.


\paragraph{Runtime Analysis}
\begin{table}[h]
\centering
\footnotesize
\begin{tabular}{lrrr}
\toprule
\textbf{Model}      & \textbf{Parameter Count} & \textbf{Time per Epoch (s)} & \textbf{Time per Inference (ms)} \\
\midrule[0.5pt]
MLP                 & 319623                  & 96 $\pm$ 4                      & 9.66                          \\
PointNet            & 154463                  & 120 $\pm$ 3                      & 11.14                          \\
GraphSAGE           & 93703                  & 147 $\pm$ 3                      & 9.74                          \\
Graph U‑Net         & 3550535                  & 1520 $\pm$ 35                       & 248.86                          \\
PointGNNConv        & 168598                  & 366 $\pm$ 4                     & 34.34                          \\
\bottomrule
\end{tabular}
\caption{Comparison of model sizes and runtimes. Inference time averaged for 200 samples.}
\label{tab:model_runtimes}
\end{table}
Table \ref{tab:model_runtimes} summarizes the inference times per sample and the parameter counts for each architecture, highlighting their relative efficiency profiles. MLP and PointNet achieve sub‑10 ms inference with strong in‑distribution accuracy (see Table \ref{tab:mse_iTest}), ideal for high‑throughput prototyping. Graph U‑Net and PointGNNConv, while costing 200–250 ms per inference, provide higher OOD fidelity (see Table \ref{tab:mse_oTest}) using advanced message passing. GraphSAGE offers a compromise with 11 ms latency and moderate generalization. 

\section{Conclusion}

We have presented AutoHood3D, the first open source, large-scale benchmark for 3D automotive hood design and one-way fluid-structure interaction. Comprising more than 16000 parametric variants and an excess of 108000 time‑resolved LES–FEA snapshots on a refined cell mesh, the dataset provides sufficient geometric diversity and high‑fidelity multiphysics solutions. Our fully documented pipeline, which spans convex shell reconstruction, cutout segmentation, 
 numerical meshing, and preCICE-mediated co-simulation, ensures complete reproducibility and extensibility to other solvers and new physical scenarios.

To support data‑driven and generative AI workflows, AutoHood3D is released in multiple modalities (STL, raw CFD/FEA meshes and fields, point clouds) and includes structured natural language prompts for supervised fine‑tuning of LLMs in text‑to‑geometry synthesis. 
We further establish performance baselines across five neural architectures: from MLP and PointNet (inference times of ~7–11 ms, in‑distribution MSE $\sim 10^{-3}$) to Graph U‑Net and PointGNNConv (34–249 ms, OOD MSE as low as $1.2 \times 10^{-2}$). These results quantify the speed-accuracy trade-offs between model classes, providing clear guidance for selecting architectures that align with specific performance and generalization requirements.

We anticipate that AutoHood3D will accelerate research in surrogate modeling, aerodynamic optimization, manufacturing process design, and physics‑informed machine learning. By lowering the barrier to large‑scale 3D FSI data generation and offering a modular open source workflow, our work lays the groundwork for future datasets to advance generative AI and data‑driven engineering.

\section{Limitations and Future Scope} \label{section:limitations}
Despite its contributions, AutoHood3D has several opportunities for improvements. First, cutouts are currently defined via 2D projections; extending them to fully 3D contours would capture more intricate geometric features. Second, our ML models are restricted to point and graph‑based architectures; investigating mesh-agnostic neural operator approaches \cite{kovachki2023neuraloperator} may improve predictive accuracy. Third, clustering cutouts by area and perimeter may overlook higher-order shape descriptors - incorporating curvature or modal deformation modes could enhance design‐space exploration. Finally, the use of MSE loss does not explicitly enforce the physical conservation of momentum or force across the fluid–structure interface; developing multiphysics‑aware training objectives that penalize these residuals could improve fidelity.

In the future, we will integrate non-linear structural solvers and introduce additional geometric parameters (e.g., hood mass distribution and material heterogeneity) to support advanced design‑optimization studies. We also plan to diversify the loading conditions, varying impulse magnitudes and orientations, to fully cover the operational envelope of manufacturing and aerodynamic scenarios.

\begin{ack}
This work was supported at the University of Michigan by Ford Motor Company under the grant ``Manufacturability-constrained closure design using physics-informed artificial intelligence". The authors thank Elliot Kimmel for discussions on FSI and Sudeep Katakol and Dr.Shivam Barwey for reviewing the initial manuscript and providing insightful feedback.
\end{ack}

\newpage
\bibliographystyle{unsrt}
\bibliography{citations}


\newpage
\newpage
\appendix
\section*{Technical Appendices and Supplementary Material}
\tableofcontents
\bigskip

\section{Hyperparameter Tuning} \label{appendix:hyperParms}
We adopt a unified set of hyperparameters that yielded stable performance across all architectures, with the specific parameters of each model (e.g., encoder-decoder configurations, hidden-channel width) tuned within this framework (see Table \ref{tab:hparams}). All models are trained using the AdamW \cite{loshchilov2017decoupled_adamW} optimizer, paired with a ReduceLROnPlateau scheduler configured as detailed in Table \ref{tab:opt_sched_hparams}. For PointGNNConv, we reduced the base learning rate to 5$\times$10$^{-5}$ to prevent training instabilities observed in early trials, and the step patience for GraphSAGE is set to 9$^*$ . Although the training pipeline supports distributed data‐parallel (DDP) execution over the full 16000+ design ensemble, the results presented here are based on a reduced subset of 4500 hood geometries trained using a single H100 GPU.

\begin{table}[h]
\centering
\footnotesize
\begin{tabular}{lccccc}
\toprule
\textbf{Weight Decay} & \textbf{Step Factor} & \textbf{Min LR} & \textbf{Step Patience} & \textbf{Epochs} & \textbf{LR Warmup} \\
\midrule
$1.00\times10^{-6}$   & 0.1             & $1.00\times10^{-7}$ & 7$^*$                & 750             & Linear, 1.5\%           \\
\bottomrule
\end{tabular}
\caption{Optimizer and scheduler hyperparameters.}
\label{tab:opt_sched_hparams}
\end{table}

\begin{table}[h]
\centering
\scriptsize
\begin{tabular}{lccccc}
\toprule
\textbf{Parameter}           & \textbf{PointGNNCon}     & \textbf{GraphUNet}        & \textbf{GraphSAGE}    & \textbf{PointNet}     & \textbf{MLP}          \\
\midrule
\texttt{encoder}             & [10,128,128,64]       & [10,128,128,64]       & [10,128,128,64]       & [10,256,128,64]       & [10,256,128,64]       \\
\texttt{decoder}             & [64,128,128,7]        & [64,128,128,7]        & [64,128,128,7]        & [64,128,256,7]        & [64,128,256,7]        \\
\texttt{nb\_hidden\_layers}  & 4                      & --                     & 4                      & --                     & 4                      \\
\texttt{size\_hidden\_layers}& 64                     & 64                     & 64                     & --                     & 256                    \\
\texttt{base\_nb}            & --                     & --                     & --                     & 8                      & --                     \\
\texttt{layer}               & --                     & SAGE                   & --                     & --                     & --                     \\
\texttt{pool}                & --                     & random                 & --                     & --                     & --                     \\
\texttt{nb\_scale}           & --                     & 5                      & --                     & --                     & --                     \\
\texttt{pool\_ratio}         & --                     & [0.5,0.5,0.5,0.5]      & --                     & --                     & --                     \\
\texttt{list\_r}             & --                     & [0.05,0.2,0.5,1,10]    & --                     & --                     & --                     \\
\texttt{batch\_size}         & 128                    & 128                    & 128                    & 128                    & 128                    \\
\texttt{nb\_epochs}          & 750                    & 352                    & 750                    & 750                    & 750                    \\
\texttt{lr}                  & 1e-4                   & 1e-4                   & 1e-4                   & 1e-4                   & 1e-4                   \\
\texttt{max\_neighbors}      & 4                      & 4                      & 4                      & --                     & --                     \\         
\texttt{r}                   & 0.05                   & 0.05                   & 0.05                   & --                     & --                     \\
\bottomrule
\end{tabular}
\caption{Hyperparameter settings for each model.}
\label{tab:hparams}
\end{table}

\subsection{Graph Construction with Max‑Neighbor Tuning} \label{appendix:graphConstruct}

In the graph model setup, each node in the mesh is connected to all neighbors within a fixed radius, defining the local geometric context, where ``max\_neighbors" (mNN) caps the number of edges per node. By building the full mesh graph once (so no contour detail is lost to sub‑sampling mesh points or repeated graph construction) and keeping the radius constant, we promote ``max\_neighbors" to the only complexity control. To measure its effect on training loss and per‑epoch runtime, we train GraphSAGE (using same settings as mentioned above except the eopchs set to 150) on the complete mesh, preserving every cutout, and systematically vary only max\_neighbors (testing 4, 8, 16, and 32), thus identifying the optimal neighborhood size that balances accuracy and speed. Here, the entire dataset is reloaded at each epoch, maintaining a constant memory footprint. This optimal setting captures critical geometric detail without incurring unnecessary computational cost.

\begin{table}[htbp]
\centering
\scriptsize
\begin{tabular}{lccccccccc}
\toprule[1.25pt]
\textbf{mNN}
  & \multicolumn{1}{c}{\begin{tabular}[c]{@{}c@{}}\textbf{$U_x$}\\\textbf{($\times10^{-2}$)}\end{tabular}}
  & \multicolumn{1}{c}{\begin{tabular}[c]{@{}c@{}}\textbf{$U_y$}\\\textbf{($\times10^{-2}$)}\end{tabular}}
  & \multicolumn{1}{c}{\begin{tabular}[c]{@{}c@{}}\textbf{$U_z$}\\\textbf{($\times10^{-2}$)}\end{tabular}}
  & \multicolumn{1}{c}{\begin{tabular}[c]{@{}c@{}}\textbf{$p$}\\\textbf{($\times10^{-2}$)}\end{tabular}}
  & \multicolumn{1}{c}{\begin{tabular}[c]{@{}c@{}}\textbf{$D_x$}\\\textbf{($\times10^{-2}$)}\end{tabular}}
  & \multicolumn{1}{c}{\begin{tabular}[c]{@{}c@{}}\textbf{$D_y$}\\\textbf{($\times10^{-2}$)}\end{tabular}}
  & \multicolumn{1}{c}{\begin{tabular}[c]{@{}c@{}}\textbf{$D_z$}\\\textbf{($\times10^{-2}$)}\end{tabular}} 
  & \multicolumn{1}{c}{\begin{tabular}[c]{@{}c@{}}{Time}\\{per Epoch}\end{tabular}}
   & \multicolumn{1}{c}{\begin{tabular}[c]{@{}c@{}}{Dataset}\\{Size (GB)}\end{tabular}}
  \\
\midrule[0.5pt]
4   &  4.65 & 1.31 &  2.00 & 4.13 & 14.56 & 3.78 & 6.56 & 147 $\pm$ 2  & 201.1 \\
8  & 3.10 & 1.19 &  1.97 & 5.15 & 9.95 & 1.76 & 3.09 & 345 $\pm$ 7 & 290.4 \\
16  &  2.93 & 1.07 &  1.71 & 4.30 & 10.49  & 4.28 & 8.53 & 443 $\pm$ 15 & 469.2 \\
32$^*$  &  1.91 &  0.98  & 1.40  & 3.35  & 8.10  & 1.51  & 2.83 & 2361 $\pm$ 245 & 826.6  \\ 
\midrule[1.25pt]
\end{tabular}
\caption{mNN Test: Mean squared error using the ID testing dataset for the different normalized fields using different mNN settings. $^*$For mNN=32 setting, the model results are reported at 130 epochs due to longer per-epoch runtime.}
\label{tab:graphCons}
\end{table}
Table \ref{tab:graphCons} demonstrates that increasing the GraphSAGE neighborhood size (mNN) steadily reduces MSE across all fields but at the expense of both runtime and data‐storage footprint. With mNN=4, the model trains in only 147 s per epoch on a 201 GB dataset, producing acceptable ID errors. Doubling to 8 neighbors halves certain displacement errors (e.g. $D_y$) but more than doubles per‐epoch time and dataset size. By mNN=32, further accuracy gains are marginal while per‐epoch time reaches $\sim$ 2361 s and the storage exceeds 800 GB, this configuration was stopped at 130 epochs due to intractability. Thus, mNN=4 delivers the most favorable speed–accuracy–size balance in this case, especially for larger graph models where high neighbor counts become prohibitively slow and memory intensive.

\section{Analysis of MLP Performance} \label{appendix:mlp}
Based on results for in-distribution tests in Sect.~\ref{sec:benchmarking}, MLP model tends to show better performance on multiple output variables (such as $U_x$, $p$) compared to more complex graph models. In this section, our aim is to explore the underlying reasons for these results, which could potentially be attributed to overfitting. Another possible explanation is that "MLPs can incorporate geometric information, given their encoder’s width being four times that of any other model." While it is true that the MLP's hidden layer width is four times the dimensionality of its encoder output—in our case, a 64 to 256 configuration—this does not necessarily imply that the MLP is aggregating information from neighboring points. Instead, each vertex's feature vector (comprising its coordinates, normals, and distance field) is processed independently through fully connected layers, without any explicit pooling or message passing over spatial neighbors. Thus, MLP’s unexpectedly strong in-distribution performance cannot be directly attributed to local geometric aggregation in the same way as with graph models. To further investigate this, we performed the following analysis: 

\begin{table}[ht]
\centering
\small
\renewcommand{\arraystretch}{1.2} 
\begin{tabular}{p{3cm} p{3.8cm} p{5.8cm} }
\toprule[1.25pt]
\textbf{Hypothesis} & \textbf{Description} & \textbf{Result} \\
\midrule[1.25pt]
A. \textbf{Non-homophily} & If adjacent nodes differ greatly, graph aggregation may harm performance. & \textbf{Rejected}: high homophily found (see Table~\ref{tab:hmply_stats}), so GNNs should benefit from local pooling. \\
\hline
B. \textbf{Sensitivity to neighbor count ($k$)} & GraphSAGE performance depends on neighborhood size. & \textbf{Confirmed}: larger $k$ reduces prediction error but at steep compute and I/O cost. \\
\hline
C. \textbf{MLP Capacity \& Overfitting} & The wide MLP memorizes training-set patterns, inflating in-distribution scores. & \textbf{Supported}: ablation shows ID error rises with hidden-dim reduction while OOD error improves (Table ~\ref{tab:mlp_oTest}). \\
\midrule[1.25pt]
\end{tabular}
\caption{Summary of hypotheses, their motivations, and empirical findings.}
\label{tab:hypotheses}
\end{table}

\subsection{Homophily Test}
The hood surface often exhibits cutout edges and curvature changes that could cause non-homophily. In such cases, neighborhood pooling of GNNs can oversmooth fine features—whereas an MLP’s point-wise mapping avoids this \cite{zhanggraph}, potentially explaining its ID test errors. Therefore, to measure the non-homophily, we computed numeric assortativity ($\bar{r}$) \cite{newman2003mixing} and a standard edge-wise multi-dimensional cosine similarity \cite{abu2019mixhop} over the entire in-distribution test data/graphs:

\begin{table}[ht]
\centering
\footnotesize
\renewcommand{\arraystretch}{1.2}
\begin{tabular}{cccccccc|c}
\midrule[1.25pt]
\textbf{Field} & $U_x$ & $U_y$ & $U_z$ & $p$ & $D_x$ & $D_y$ & $D_z$ & \textbf{Multi-dim Cosine} \\
\hline
\textbf{$\bar{r}$} & +0.9756 & +0.9739 & +0.9711 & +0.9650 & +0.9910 & +0.9933 & +0.9877 & +0.9449 \\
\hline
\textbf{$\sigma_r$} & 0.0037 & 0.0106 & 0.0059 & 0.0058 & 0.0301 & 0.0154 & 0.0322 & 0.0109 \\
\midrule[1.25pt]
\end{tabular}
\caption{Mean assortativity ($\bar{r}$) and its standard deviation for each physical field with cosine similarity.}
\label{tab:hmply_stats}
\end{table}

Interpretation: All fields exhibit strong homophily ($\bar{r}$ > 0.96), so neighbor aggregation should be beneficial for GNNs—not detrimental—if effectively leveraged. This can be done by carefully selecting the "max neighbors" (mNN), as the quality of feature aggregation in GNNs is significantly influenced by this choice. This is the focus of Sec.~\ref{appendix:graphConstruct}, where we explore the impact of mNN selection on GNN performance in more detail.

\subsection{GraphSAGE mNN Sweep}
In a separate experiment (see Table~\ref{tab:graphCons}), we varied GraphSAGE’s neighbor count (mNN = 4,8,16,32). Increasing the neighborhood size from 4 to 32 reduces ID MSE by up to 40-50\%, but:\\
Compute Time: per-epoch time grows by an order of magnitude.\\
Data I/O: dataset size inflates proportionally to neighborhood features stored.\\
Higher mNN reduced in-distribution errors at the expense of increased computation time, illustrating the classic trade-off of neighborhood aggregation.

\subsection{MLP Hidden-Size Ablation Study}
Our ablation study compared MLPs with hidden‐layer widths of 256, 128 and 64—keeping all other hyperparameters consistent with Table~\ref{tab:hparams}. The following is the complete set of results:

\begin{table}[!htp]
\centering
\footnotesize 
\begin{tabular}{lccccccc}
\toprule[1.25pt]
\textbf{Model}
  & \multicolumn{1}{c}{\begin{tabular}[c]{@{}c@{}}\textbf{$U_x$}\\\textbf{($\times10^{-2}$)}\end{tabular}}
  & \multicolumn{1}{c}{\begin{tabular}[c]{@{}c@{}}\textbf{$U_y$}\\\textbf{($\times10^{-2}$)}\end{tabular}}
  & \multicolumn{1}{c}{\begin{tabular}[c]{@{}c@{}}\textbf{$U_z$}\\\textbf{($\times10^{-2}$)}\end{tabular}}
  & \multicolumn{1}{c}{\begin{tabular}[c]{@{}c@{}}\textbf{$p$}\\\textbf{($\times10^{-2}$)}\end{tabular}}
  & \multicolumn{1}{c}{\begin{tabular}[c]{@{}c@{}}\textbf{$D_x$}\\\textbf{($\times10^{-2}$)}\end{tabular}}
  & \multicolumn{1}{c}{\begin{tabular}[c]{@{}c@{}}\textbf{$D_y$}\\\textbf{($\times10^{-2}$)}\end{tabular}}
  & \multicolumn{1}{c}{\begin{tabular}[c]{@{}c@{}}\textbf{$D_z$}\\\textbf{($\times10^{-2}$)}\end{tabular}} \\
\midrule[0.5pt]
MLP 256 & 0.25 & 0.27 & 0.44 & 0.37 & 2.80 & 0.51 & 0.76 \\
MLP 128 & 0.32 & 0.30 & 0.49 & 0.47 & 4.59 & 0.75 & 0.91 \\
MLP 064 & 0.39 & 0.37 & 0.53 & 0.58 & 7.63 & 0.74 & 1.02 \\
\midrule[1.25pt]
\end{tabular}
\caption{ID Test: Mean squared error on the different normalized fields for all the MLP models.}
\label{tab:mlp_iTest}
\end{table}

\begin{table}[!htp]
\centering
\footnotesize 
\begin{tabular}{lccccccc}
\toprule[1.25pt]
\textbf{Model}
  & \multicolumn{1}{c}{\begin{tabular}[c]{@{}c@{}}\textbf{$U_x$}\\\textbf{($\times10^{-2}$)}\end{tabular}}
  & \multicolumn{1}{c}{\begin{tabular}[c]{@{}c@{}}\textbf{$U_y$}\\\textbf{($\times10^{-2}$)}\end{tabular}}
  & \multicolumn{1}{c}{\begin{tabular}[c]{@{}c@{}}\textbf{$U_z$}\\\textbf{($\times10^{-2}$)}\end{tabular}}
  & \multicolumn{1}{c}{\begin{tabular}[c]{@{}c@{}}\textbf{$p$}\\\textbf{($\times10^{-2}$)}\end{tabular}}
  & \multicolumn{1}{c}{\begin{tabular}[c]{@{}c@{}}\textbf{$D_x$}\\\textbf{($\times10^{-2}$)}\end{tabular}}
  & \multicolumn{1}{c}{\begin{tabular}[c]{@{}c@{}}\textbf{$D_y$}\\\textbf{($\times10^{-2}$)}\end{tabular}}
  & \multicolumn{1}{c}{\begin{tabular}[c]{@{}c@{}}\textbf{$D_z$}\\\textbf{($\times10^{-2}$)}\end{tabular}} \\
\midrule[0.5pt]
MLP 256 & 1.49 & 1.89 & 2.97 & 4.96 & 181.02 & 5.87 & 19.81 \\
MLP 128 & 1.58 & 1.60 & 2.68 & 4.65 & 175.28 & 7.58 & 22.91 \\
MLP 064 & 2.14 & 1.09 & 1.59 & 3.38 & 124.11 & 8.50 & 20.55 \\
\midrule[1.25pt]
\end{tabular}
\caption{OOD Test: Mean squared error on the different normalized fields for all the MLP models.}
\label{tab:mlp_oTest}
\end{table}

ID Trend: error increases steeply as hidden dimension shrinks, confirming that the 256-dim model leverages high capacity to fit training data.\\
OOD Trend: the smaller MLPs generalize marginally better, indicating the largest network is more susceptible to overfitting the data.\\
Together, these studies suggest (Table~\ref{tab:hypotheses}) that the advantage of MLPs on ID tests is not directly due to hidden neighbor aggregation; rather, their wider layers facilitate point-wise memorization of global patterns. On the other hand, graph-based models take advantage of the existing homophily within the data, but they must strike a careful balance between neighborhood size and the risk of oversmoothing, as well as the associated computational overhead. As such, the choice of model for practical surrogate design will ultimately depend on the specific application: MLPs are better suited for high-throughput, in-distribution tasks, while GNNs—when tuned for maximum neighborhood size (mNN)—are more effective for robust generalization, assuming sufficient computational resources are available.

\section{Numerical Solvers} \label{appendix:solvers}
The following subsections detail the configuration of each solver and the corresponding test cases. 
\subsection{Fluid Solver}
UM\_pimpleFoam is a CPU-based solver built on OpenFOAM’s pimpleFoam solver. Unlike the standard pimpleFoam solver, UM\_pimpleFoam incorporates an additional transient acceleration source term in the momentum‐conservation equation to capture the inertial body force arising from the prescribed fluid acceleration. For this study, LES simulation is conducted with the Spalart-Allmaras Delayed Detached Eddy Simulation (DDES) model \cite{spalart2006new}. UM\_pimplefoam solves the incompressible Navier-Stokes equations on a nonuniform grid. Following are the system of equations along with the additional acceleration term ($A_i$). The mass-conservation equation is:
\begin{equation}
     \frac{\partial u_i}{\partial x_i} = 0, 
\end{equation}
where $x_i$ = (x, y, z) are the Cartesian coordinates, and the components of the velocity vector are $u_i$ = (u, v, w). The fluid is treated as incompressible with constant density $\rho$, therefore the momentum equation is normalized by $\rho$ \cite{ferziger2002computational}. The resulting form is:
\begin{equation}
\underbrace{\frac{\partial u_i}{\partial t}}_{\text {local (unsteady) acceleration }}+\underbrace{u_j \frac{\partial u_i}{\partial x_j}}_{\text {convective momentum transport }}=\underbrace{-\frac{1}{\rho} \frac{\partial p}{\partial x_i}}_{\text {pressure-gradient force }}+\underbrace{\nu \frac{\partial^2 u_i}{\partial x_j \partial x_j}}_{\text {viscous diffusion }}+A_i  
\end{equation}
Here, the left‐hand side collects the inertial terms, while the right‐hand side comprises pressure, viscous, and body‐force contributions. The term $A_i$	
explicitly represents the time-varying acceleration applied to each fluid cell. Therefore, due to the incompressible formulation, the fluid density $\rho$ is implicitly incorporated through kinematic viscosity ($\nu = \mu / \rho$).


\subsection{Solid Solver}
UM\_solidDisplacementFoam is a CPU-based solver built on OpenFOAM’s solidDisplacementFoam. We chose OpenFOAM’s structural solver for its seamless integration with the fluid solver and streamlined mesh generation, since the solid mesh derives directly as the complementary subset of the fluid-domain mesh. The solver uses a small-strain elastic deformation model to solve the stress and displacement of the hood, and solves the stress-strain relationship defined by Hooke’s law. Since the deformations during the dip process are not large \cite{kim2015predictionHood, kim2014microhood}, the linearity assumption between stress and strain remains valid. However, for larger deformations, different solvers can be incorporated using the preCICE adapter \cite{preCICEv2}. Unlike the standard implementation, our solver embeds the force patch calculation routine \cite{cardiff-2016SolidFoam} directly within its code, eliminating the need for any external function. 
Following equation is the transient linear momentum balance for a deformable solid \cite{fenner1986engineeringSolid}:
\begin{equation}
\rho\,\frac{\partial^2 D_i}{\partial t^2} \  \ =
\underbrace{\frac{\partial \sigma_{ij}}{\partial x_j}}_{\text{divergence of stress}}
\;+\;
\underbrace{b_i}_{\text{body‐force density}},
\end{equation}
where $D_i$ are the displacement vector components and $\sigma_{ij}$ is the  Cauchy stress. The small‐strain tensor is defined as:
\begin{equation}
\varepsilon_{ij} = \underbrace{\frac{1}{2}\Bigl(\frac{\partial D_i}{\partial x_j}+\frac{\partial D_j}{\partial x_i}\Bigr)}_{\text{small‐strain tensor}}.
\end{equation}
And the Hooke’s law for a linear elastic, isotropic material:
\begin{equation}
\sigma_{ij}
=
\underbrace{\lambda\,\varepsilon_{kk}\,\delta_{ij}}_{\text{volumetric response}}
\;+\;
\underbrace{2\,\mu\,\varepsilon_{ij}}_{\text{shear response}}, 
\end{equation}
expressing the stress tensor ($\sigma_{ij}$) in terms of volumetric response (through Lamé’s first parameter $\lambda$) and shear response (through the shear modulus $\mu$) \cite{slone-2002Solid}. Thermal stresses are neglected in our formulation, and the corresponding terms are therefore omitted from the governing equations. The deformation of the solid region is solved based on the pressure distribution at the interface for the previous timestep. 
\paragraph{Extending FEA} Leveraging preCICE’s \cite{preCICEv2} solver‑agnostic coupling, our FSI framework can be extended to incorporate non-linear structural solvers, native to OpenFOAM or through external libraries such as deal.II \cite{bangerth2007deal2} and FEniCS \cite{alnaes2015fenics}, facilitating multiphysics analyses across various engineering domains and requirements.

\subsection{Solver Validation}
Following validation cases are presented to support the simulation settings and to justify the selection of parameters such as mesh refinement levels and solver specifications. The cases are provided along with the associated codes and workflows.
\subsubsection{Fluid Test Case}
\begin{figure} [!htb]
  \centering
  \includegraphics[width=1.0\linewidth]{ 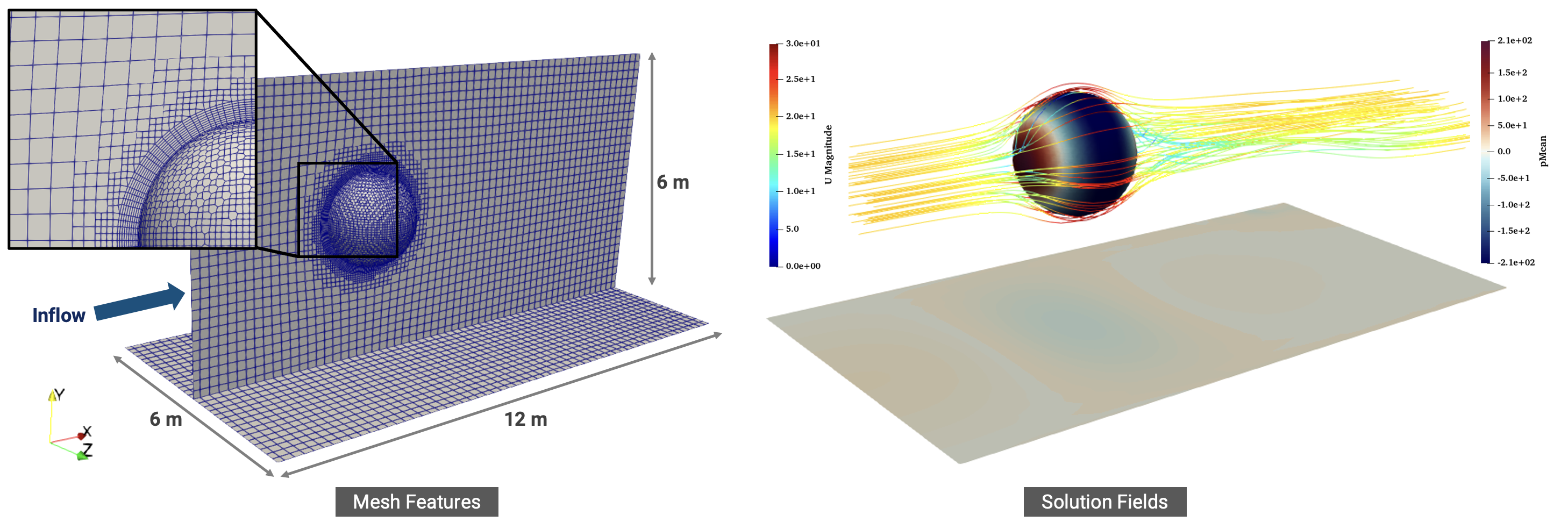}
  \caption{Computational setup and results for flow over a sphere. (Left) Mesh features showing domain dimensions, refinement near the sphere, and inflow direction. (Right) Solution fields illustrating velocity streamlines colored by velocity magnitude, and pressure contours on the sphere and ground plane, highlighting the pressure distribution.}
  \label{fig:fluid_val}
\end{figure}

This case considers the simulation of a 3D sphere placed in a domain of dimensions 12$\times$6$\times$6 m, with a uniform flow field to evaluate the pressure drag force, as shown in Fig.\ref{fig:fluid_val}. Inflow consists of atmospheric air at a velocity of $U$ = 20 m/s and a temperature $T$ = 293 K, corresponding to a density $\rho$ = 1.21041 kg/m$^3$. The sphere has a radius of 1 m, resulting in a frontal area of A = 3.14 m$^2$. Based on flow conditions, the Reynolds number is calculated as $Re$ = 2.47$\times$10$^6$ , placing it well within the turbulent regime where the drag coefficient C$_d$ for a smooth sphere is approximately 0.47 (based on \cite{schiller1933drag}). The computational mesh is constructed with a characteristic grid spacing of $\Delta x \simeq$ 0.039 m to ensure adequate resolution of the flow features around the sphere. The analytical force can be computed as $F_d = \frac{1}{2} \, \rho \, u^2 \, C_d \, A$, and using the provided values, it is found to be 355.582 N. Using the simulation results, the drag force is calculated as $F^{Sim}_d = \sum (p \times \vec{n} ) \rho A$, which yields 357.049~N, corresponding to an error of 0.41\%. The simulation accurately predicts the drag force on the sphere, with a computed error of only 0.41\% compared to the analytical value. This close agreement validates the mesh quality and the overall solver setup. 

\subsubsection{Solid Test Case}
\begin{figure} [!htb]
  \centering
  \includegraphics[width=1.0\linewidth]{ 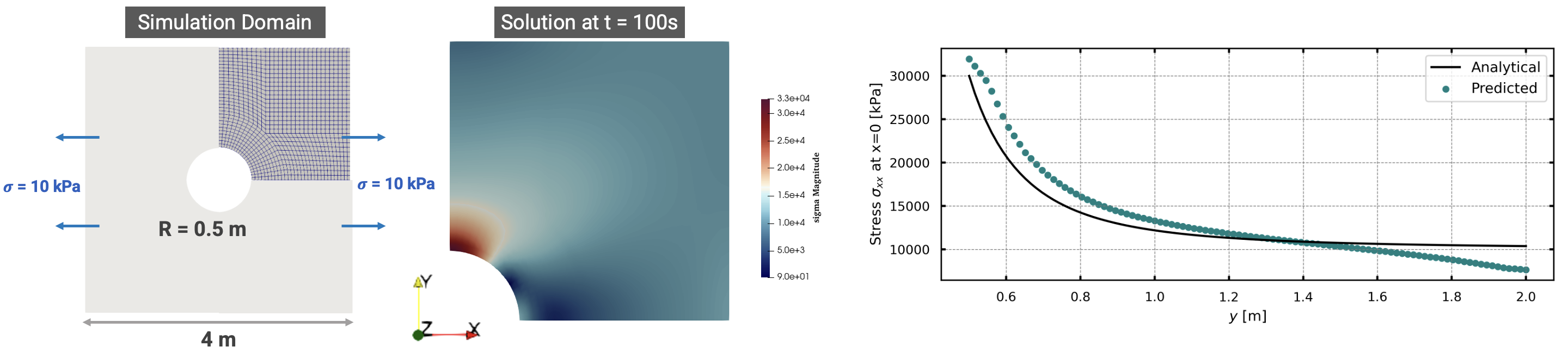}
  \caption{Computational setup and results for time varying flow over a flap. (Left) Mesh features showing domain dimensions, refinement near the hole, and traction directions. Solution fields illustrating stress contour on the plate at time $t = 100s$ (middle) and normal stress along the vertical symmetry plotted on the right.}
  \label{fig:solid_val}
\end{figure}

To test the solid solver, a 2D problem focused on steady-state linear elastic stress analysis on a square plate with a circular hole at its center is used. The dimensions of the plate are: side length 4 m and radius R = 0.5 m. It is loaded with a uniform traction of $\sigma$ = 10 kPa on its left and right faces, as shown in Figure \ref{fig:solid_val}. Due to dual symmetry, only a quarter of the domain is meshed. We adopt a plane‐stress formulation, neglecting all out‐of‐plane stress and strain components. For an infinitely thin plate with a central circular hole under remote uniaxial tension, there is an exact analytical solution. In particular, the normal stress on the symmetry plane is given by \begin{equation}
\left(\sigma_{x x}\right)_{x=0}= \begin{cases}\sigma\left(1+\frac{R^2}{2 y^2}+\frac{3 R^4}{2 y^4}\right) & \text { for }|y| \geq R \\ 0 & \text { for }|y|<R\end{cases}
\end{equation}
The middle plot in Figure \ref{fig:solid_val} presents the von Mises stress ($|\sigma|$) field at t=100 s on the quarter‑domain mesh, illustrating the stress concentration around the circular hole under a uniform 10 kPa tensile load. Along the symmetry line x=0, the computed normal stress is plotted against the analytical solution derived above. Green markers represent numerical predictions, which closely follow the solid black curve of the analytical solution, with slight overprediction near the hole rim and good agreement farther away.

\subsubsection{FSI Test Case}
\begin{figure} [!htb]
  \centering
  \includegraphics[width=1.0\linewidth]{ 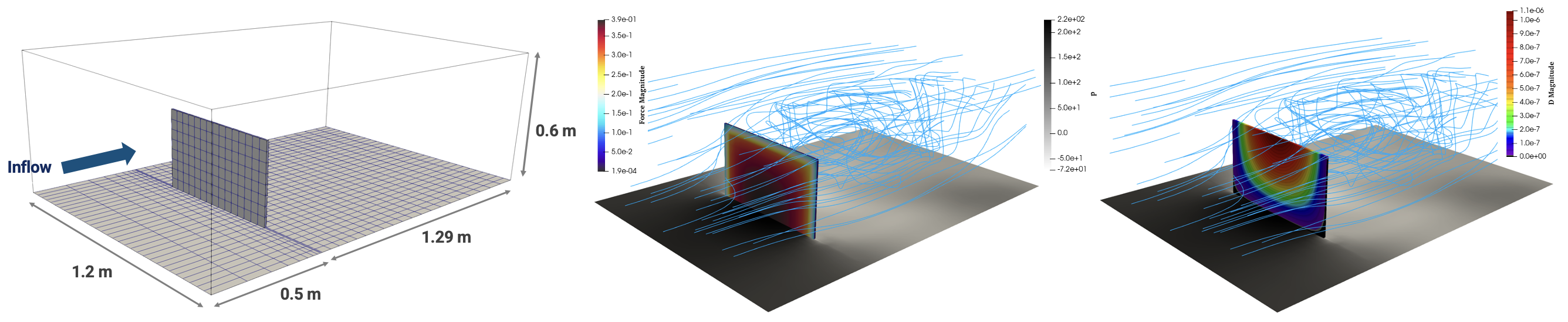}
  \caption{Computational setup and results for time varying flow over a flap. (Left) Mesh features showing domain dimensions, refinement near the flap, and inflow direction. Solution fields illustrating velocity streamlines, and pressure contours on the ground plane with force contour (middle) and displacement contour (right) on the flap.}
  \label{fig:fsi_val}
\end{figure}

For the 3D FSI test case, the simulation consists of a thin 3D flap anchored at the bottom in a channel with a time-varying inflow from the left, as shown in Fig. \ref{fig:fsi_val}. One-way coupling is used between the fluid and the solid solvers as described in the previous sections. The overall dimensions of the domain are 1.8 $\times$ 0.6 $\times$ 1.2 m with a prescribed no slip ($u=0$) condition in the top and bottom channel walls, with inflow velocity described as $u$ = $(10,0,0)$. 
The case is adapted from \cite{schott2019monolithic}, with the properties of the selected material set as: density ($\rho_s$) = 2700 kg/m$^3$, Poison ratio ($\nu$) = 0.33 and Young’s modulus ($E$) = 68.9e9 N/m. The flap has dimensions of 0.01 $\times$ 0.35 $\times$ 0.6 m and clamped to the bottom. Three different levels of mesh refinements are simulated (cases 1–3 employ successive 2$\times$ mesh refinements), with the fluid solver settings adapted from the previous section. The force on the flap is measured at the location [0.5,0.35,0.0].
\begin{figure} [!htb]
  \centering
  \includegraphics[width=0.90\linewidth]{ 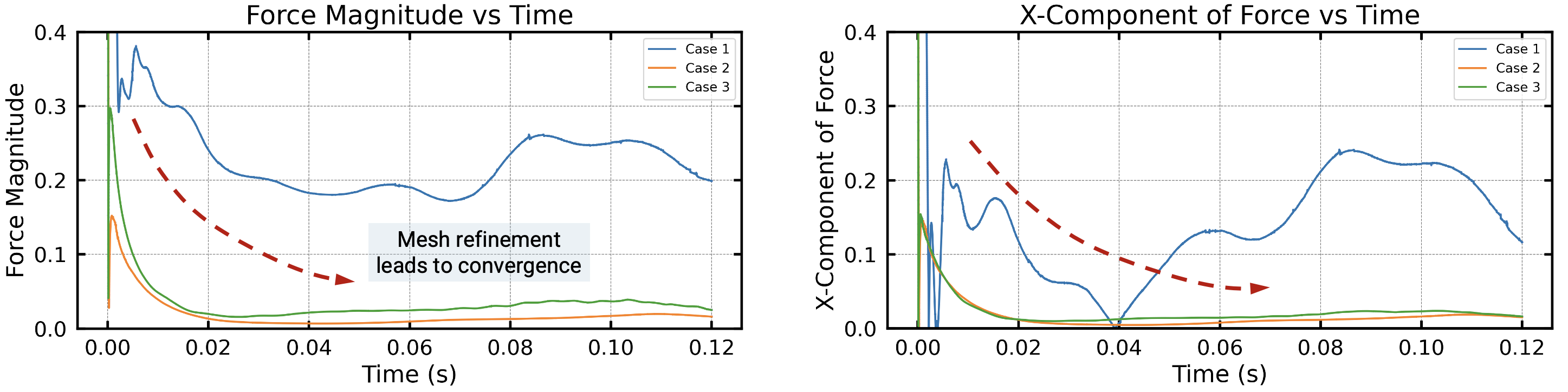}
  \caption{Force response at a flap location under uniform loading for three mesh refinements: Case 1 (baseline), Case 2 (2× refinement), and Case 3 (4× refinement). Left: total force magnitude vs. time. Right: X‐component of force vs. time. The red dashed arrow highlights convergence toward a smooth solution as mesh density increases.}
  \label{fig:fsi_force}
\end{figure}
In Figure \ref{fig:fsi_force}, the force time histories for the two finer meshes (Cases 2 and 3) nearly coincide, demonstrating convergence as the mesh resolution doubles twice. In contrast, the coarsest mesh (case 1) exhibits pronounced oscillations, particularly in the X-component trace, due to under-resolved pressure gradients and numerical interpolation noise across the flap interface. As resolution increases, these spurious fluctuations decrease, and the solution smoothly decays to its steady‐state value.

\section{Dataset} \label{appendix:dataset}

\subsection{Dataset Structure and URLs} \label{sec:links}

\begin{itemize}
    \item Code available at : \url{https://github.com/vanshs1/AutoHood3D/}

    \item Dataset of SFT LLM prompts (in 06\_LLM\_Generation/Final\_consolidated\_prompts.jsonl): \url{https://github.com/vanshs1/AutoHood3D/}

    \item Dataset of hood base shells/skins: \url{https://doi.org/10.7910/DVN/9268BB}
    
    \item Dataset of 4k hoods 
    \begin{itemize}
        \item STLs: \url{https://doi.org/10.7910/DVN/HEILMB}
        \item Simulation Data (raw): \url{https://doi.org/10.7910/DVN/VCKOK5}
        \item Processed for ML task: \url{https://doi.org/10.7910/DVN/6OAFF8}
        \item Processed for ML task (Graphs): \url{https://doi.org/10.7910/DVN/WODNWY}
        \item Test ML workflow$^{**}$: \url{https://doi.org/10.7910/DVN/FSYRJA}
    \end{itemize}
    
    \item Dataset of 12k hoods 
    \begin{itemize}
        \item STLs: \url{https://doi.org/10.7910/DVN/Z0VXLI}
    \end{itemize}

    \item Dataset of 12k hoods (randomized)
    \begin{itemize}
        \item STLs: \url{https://doi.org/10.7910/DVN/OJXIS1}
    \end{itemize}

\end{itemize}
The updated URLs for the datasets, particularly the simulation data for the 12k case due to its large size, and the corresponding Croissant metadata for each dataset are shared on the GitHub repository. A compact ML test set$^{**}$ of 100 preprocessed cases is included for the rapid validation of training workflows. From these, users can easily sample smaller subsets (e.g., 12 or 32 batch sizes) to perform local testing of the ML pipelines.

\subsection{Contents}
The dataset contents are as follows:
\begin{itemize}
    \item 10000+ randomized hoods cut from 108 unique base hood shells.
    \item 12000+ clustered hoods cut from 108 unique base hood shells.
    \item An additional 4500+ clustered hoods.
    \item JSON files for 2500+ prompts.
    \item 108 base shells for making new geometries.
    \item 1750 different curve cut-out files.
\end{itemize}

\begin{table}[htbp]
    \centering
    \renewcommand{\arraystretch}{1.3}
    \begin{tabular} {|p{4cm}|p{9.5cm}|}
        \hline
        \textbf{File Format} &  \textbf{Description} \\
        \hline
        \multicolumn{2}{|c|}{ \textit{Hood geometries} } \\
        \hline
        geo\_***\_clusterID\_x\_...stl & Surface mesh of the hood geometry from hood of base shell ***, cluster x (see Fig.\ref{fig:geo_features})\\
        \hline
        geo\_***\_...stl & Surface mesh of the hood geometry from hood of base shell *** with random cutouts (no cluster sampling) (see Fig.\ref{fig:geo_features})\\
        \hline
        \multicolumn{2}{|c|}{ \textit{Results and other data} } \\
        \hline
        solid (or fluid)\_...vtp & Raw CFD and FEA data, contains field values for the physical quantities. VTP is a poly-data format derived from VTK data type.\\
        \hline
        batch\_**.pt & Torch dataset format containing fields for ML training. \\
        \hline
        x.json & JSON file containing LLM prompts or structured geometry list for creating ML training dataset. \\
        \hline
    \end{tabular}
    \caption{Summary of the dataset contents}
    \label{tab:dataset_config_multi}
\end{table}

\begin{figure}[h]
  \centering
  \includegraphics[width=0.90\linewidth]{ 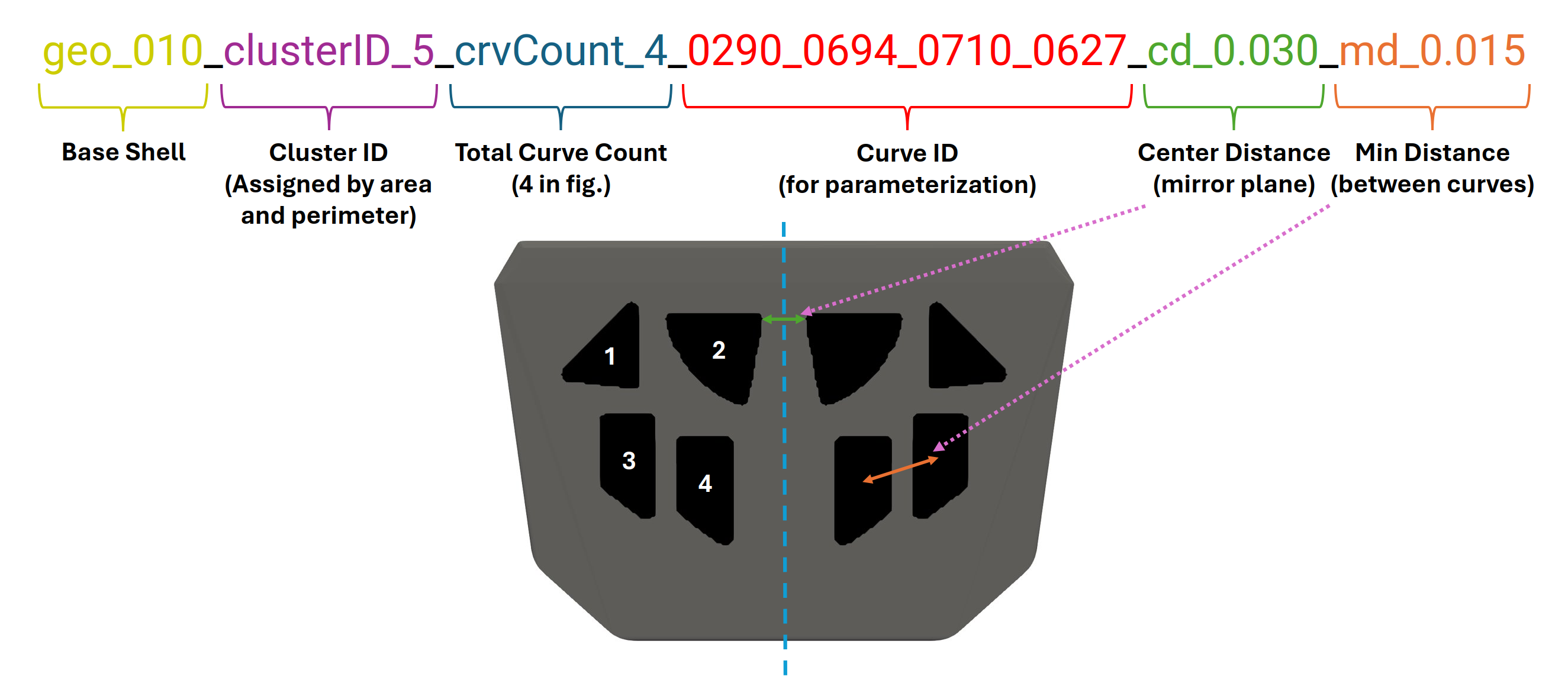}
  \caption{Example hood showing the physical definitions of the geometry naming convention}
  \label{fig:geo_features}
\end{figure}

In Figure \ref{fig:geo_features}, the naming convention of a standard clustered geometry is shown. The base shell geo\_010 refers to the standard outer shell from which the cutouts are carved. The ``clusterID" 5 is assigned based on the strata of the perimeter and the area of the cuts. Further details on clustering are provided in \ref{appendix:clustering}. All geometries are symmetrical, with ``crvCount" referring to the number of cutouts on either side of the hood. ``Curve ID" 0290\_0694\_0710\_0627 functions to parameterize the geometry. The center distance ``cd \_0.030" refers to the minimum distance between the cutouts across the mirror plane, and the ``md\_0.015" refers to the minimum distance between the center of two cutouts on either side of the hood.

\subsection{Hood Geometry Variations} \label{appendix:hoodVariations}
Starting from a single inner‑hood CAD model, we leverage a suite of Python scripts built on pyMadCAD \cite{pymadcad} to automatically generate the full exterior shell and project engineered features onto a reference plane. This end‑to‑end, workflow transforms one base geometry into numerous distinct hood variants, capturing a wide spectrum of cutout topologies and enabling rich exploration of design and fluid–structure responses. The detailed steps are as follows:

\begin{itemize}
  \item \textbf{Convex‐Hull Shell Reconstruction:} Starting from the inner hood mesh, compute its convex hull to form an outer envelope that closes all recesses. Extrude a uniform offset (e.g. 10 mm) normal to the hull surface to produce the complete 3D hood shell.
  \item \textbf{Planar Projection:} Translate and rotate the shell so that its engineered features face a reference plane perpendicular to the inlet flow. Project the 3D surface orthogonally onto this plane to obtain a binary silhouette highlighting potential cutout regions.
  \item \textbf{Segmentation and Mask Extraction:} Apply the SAM‑2 \cite{ravi2024sam2} segmentation model to the 2D projection to isolate individual cutout masks. Post‑process masks to remove noise and enforce connectivity, ensuring that each mask corresponds to a single-engineered opening.
  \item \textbf{Boundary Sampling and Curve Database:} Trace the contour of each mask and uniformly sample boundary points to generate an ordered point‐cloud curve. Store these curves in a parameter database for subsequent re‐embedding.
  \item \textbf{Cutout Reintegration:} Map selected boundary curves back onto the 3D shell surface and subtract them to carve precise openings. Enforce symmetry and user‐defined spacing constraints to create the final engineered hood variations.
\end{itemize}
\begin{figure}[h]
  \centering
  \includegraphics[width=1.0\linewidth]{ 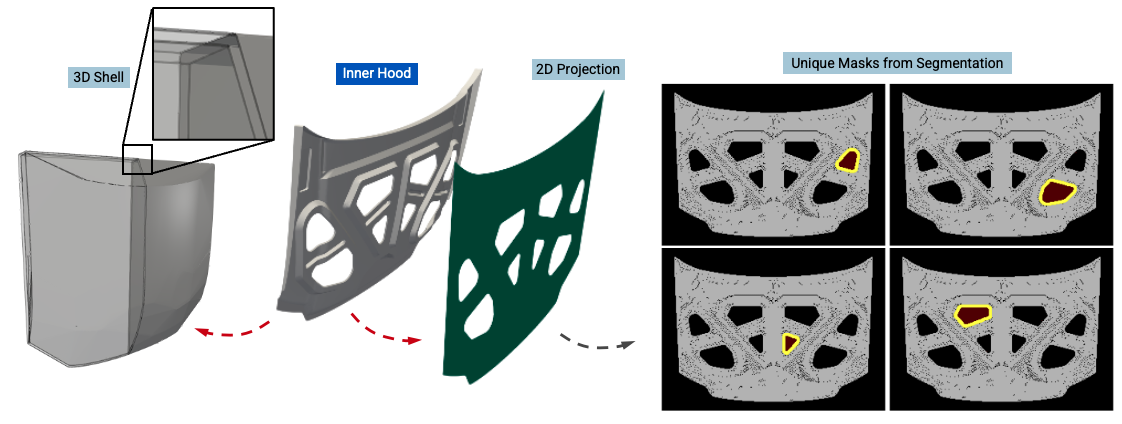}
  \caption{Generation of engineered hood shells from inner‑hood CAD. The process begins with the inner‑hood surface (second panel), which is convex‑hulled and offset to form the complete 3D shell (first panel). This shell is projected onto a 2D plane (third panel) and segmented via SAM-2 \cite{ravi2024sam2} to extract unique cutout masks (right panels), each of which defines a boundary curve used to parameterize and re‑embed the engineered openings into the final shell geometry.}
  \label{fig:projection}
\end{figure}

\subsection{Clustering Cutouts} \label{appendix:clustering}
\begin{figure}[h]
  \centering
  \includegraphics[width=1.0\linewidth]{ 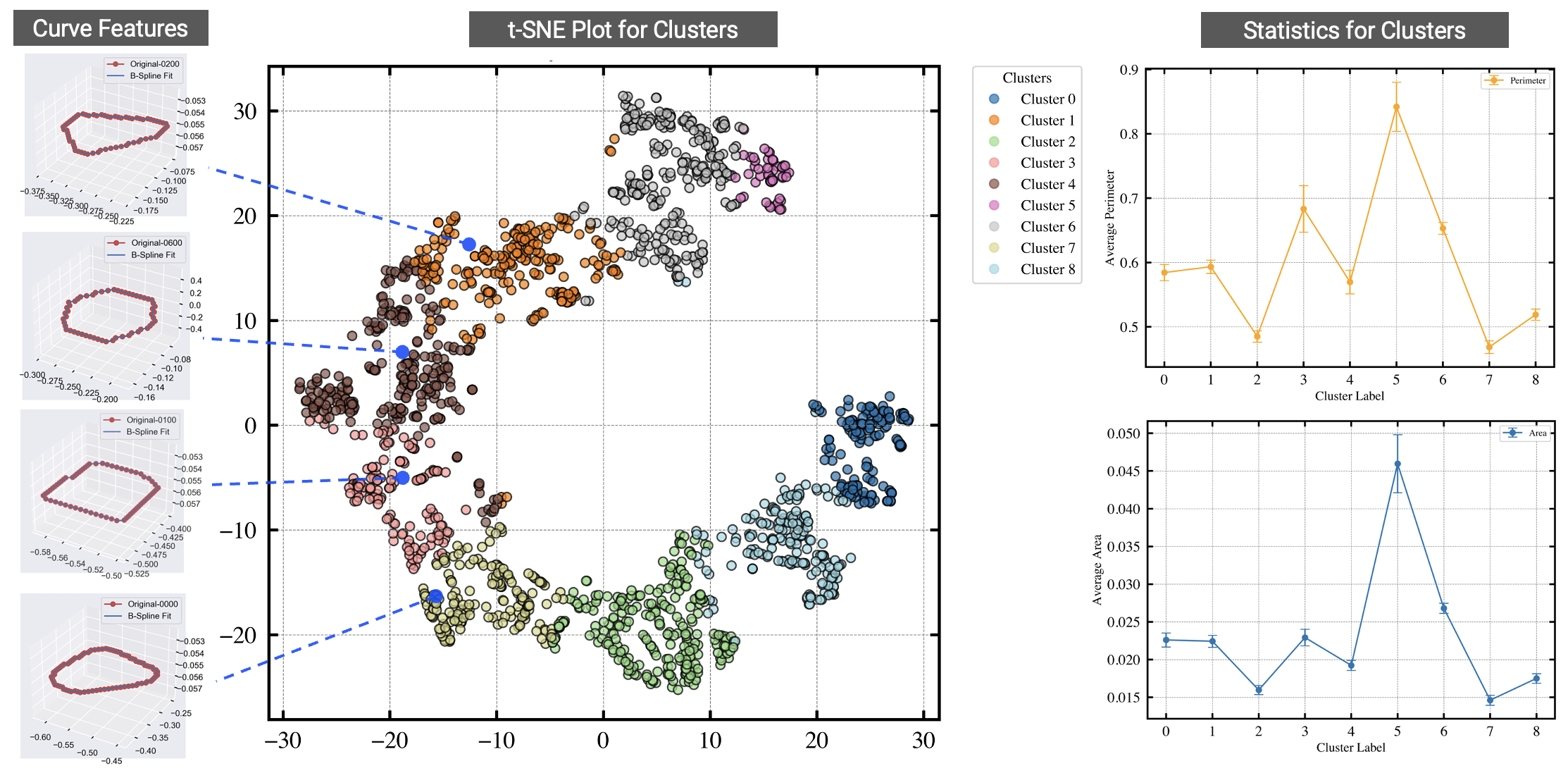}
  \caption{Clustering of extracted cutout curves. Left: representative boundary‐curve features (original and B‑spline fits) for four sample clusters. Center: t‑SNE projection of all curves, colored by cluster label (0–8), illustrating distinct groupings based on perimeter and area metrics. Right: mean perimeter (top) and area (bottom) with error bars for each cluster, highlighting the geometric diversity captured across the nine classes.}
  \label{fig:clusters}
\end{figure}
Figure \ref{fig:clusters} delineates nine clusters of engineered cutouts stratified by perimeter and area. Cluster 5 comprises the most elaborate shapes, with a mean perimeter of approximately 0.84 and mean area around 0.045. In contrast, clusters 2 and 7 represent the most compact designs, exhibiting mean perimeters between 0.43 and 0.49 and mean areas near 0.015. In contrast, several intermediate clusters (e.g., Clusters 1, 3, and 4) exhibit partial overlap, indicating gradual transitions in perimeter and area metrics and demonstrating that certain mid‑range shapes share similar feature compositions. This overlap underscores the continuous nature of our design manifold and suggests opportunities for interpolation between cluster centroids when generating new variants. This calibrated partitioning of cutout topologies provides the dataset with a comprehensive representation of feature scales and intricacies, supporting systematic investigation of geometry‐driven variations in fluid–structure response.



\subsection{Automated Simulations}
We developed a suite of six interconnected Python scripts to fully automate the end‑to‑end HPC workflow. A primary ``manager" script handles mesh generation, solver invocation, and directory I/O for each design. Two ``replicator" scripts clone and distribute case directories across compute pools. Two SLURM \cite{yoo2003slurm} utilities generate and submit batch jobs for each case. Finally, a top‑level orchestration script monitors node availability, cleans temporary files, assigns pools to nodes, and invokes the subordinate scripts in sequence.

\subsection{Analysis of Simulation Results} \label{appendix:analysisSims}
\begin{figure}[h]
  \centering
  \includegraphics[width=1.0\linewidth]{ 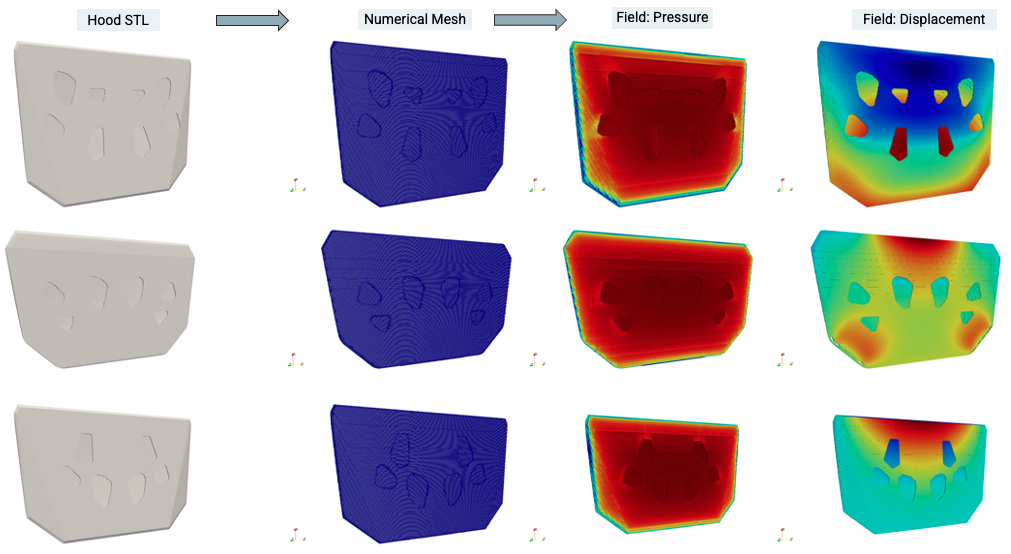}
  \caption{Representative hood variants (left) with corresponding CFD meshes (second column), pressure distributions (third column), and structural displacement fields (right column) for three sample geometries}
  \label{fig:sols}
\end{figure}

Each row in Figure \ref{fig:sols} illustrates how distinct cutout geometries yield different deformation patterns under identical loading. As the cutout topology varies across the three samples, so does the displacement pattern despite similar looking pressure fields. The large central openings of the top design drive pronounced edge deflections, the smaller scattered apertures of the middle design produce localized deformation around each edge, and the uniformly distributed holes of the bottom variant produce a more evenly distributed displacement profile in the lower half but a higher concentrated deflection on the top. Next, we visualize the simulation results for all the 4k dataset to characterize the global variation in pressure and deformation.

\begin{figure}[h]
  \centering
  \includegraphics[width=1.0\linewidth]{ 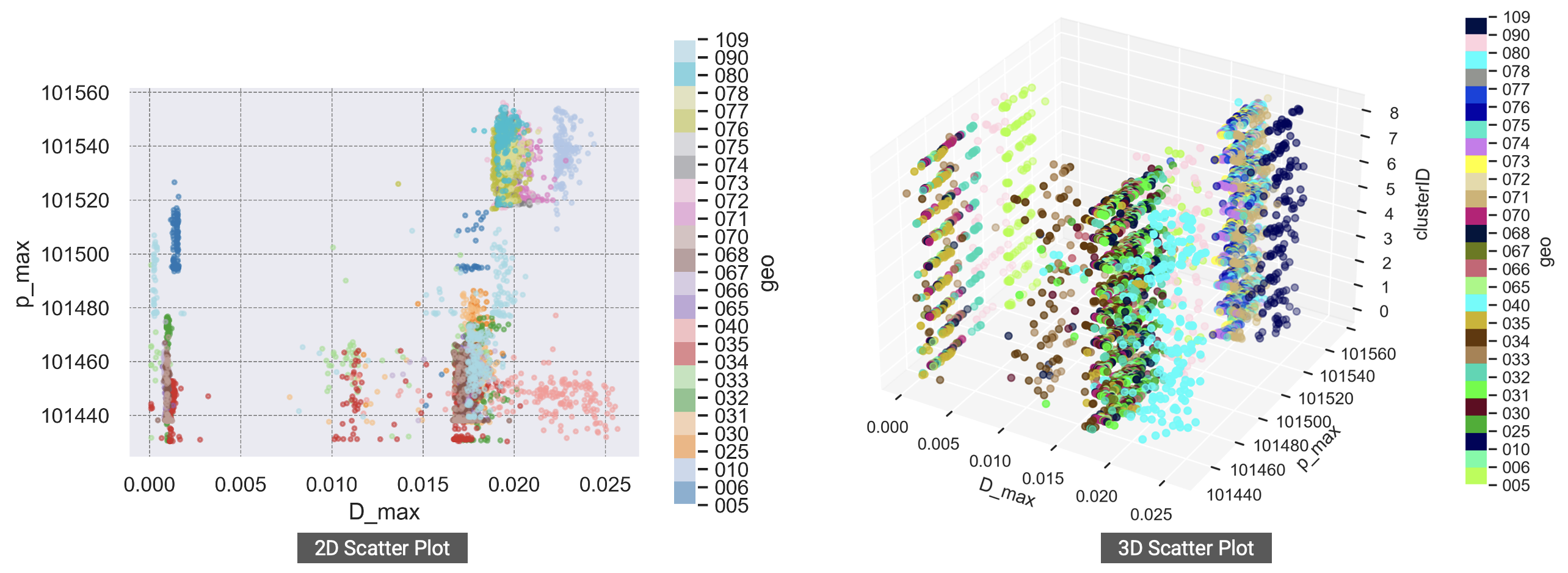}
  \caption{Raw CFD results: maximum pressure $p_{max}$ (Pa) vs. maximum deflection  $D_{max}$ (m). Left: colored by geometry index; right: stratified by cluster ID.}
  \label{fig:cfdScatter}
\end{figure}

The plots in Figure \ref{fig:cfdScatter} illustrate how geometric variability drives coupled pressure–deflection responses across the dataset. In the 2D scatter (left), each point’s color encodes its base‑geometry index, revealing that certain designs, such as ``geo 0", consistently exhibit low $D_{max}$ even under moderate $p_{max}$, whereas others (e.g., ``geo 7") manifest higher deflection at similar pressures. The 3D plot (right) layers these same data by cluster ID, showing that extremes in both deflection and pressure occur predominantly in Cluster 5 (IDs around 070–090). Within each cluster band, the spread of geometry‑indexed colors demonstrates that individual hood shapes produce unique pressure–deflection signatures. This interplay between cluster‑level typologies and specific geometric variations highlights the sensitivity of the structural response to cutout details and underscores the importance of a thorough design exploration.

\subsection{LLM Prompt}
We designed a structured meta‑prompt to guide Gemma3’s vision‑LLM into producing paired text–point‑cloud training examples. For each hood image and the corresponding variation specification, the prompt instructs the model to generate: (1) a User Input section that formalizes the design request from the base description and variation details; (2) a Chain-of-thought section that explicitly reasons through curve counts, spacing, symmetry, and surface orientation; and (3) a Solution placeholder for the resulting point‑cloud output. By enforcing this three‑part schema, we automatically extract self‑documented examples, complete with transparent intermediate reasoning, suitable for supervised fine‑tuning of downstream LLMs on CAD text‑to‑geometry tasks. The code along with necessary details is shared online and the prompt is shown below. 
\begin{tcbraster}[raster valign=top, raster halign=center,raster columns=1, raster rows=1, colframe=white,colback=white,colbacktitle=teal!50!white,] \label{block:dataset}

\begin{tcolorbox}[title = Meta Prompt,colback=black!5!white,colframe=teal!75!black,fonttitle=\bfseries, width = 0.475\linewidt]
\footnotesize{
You are a CAD‐focused Vision‐LLM. Analyze the image provided along with these guidelines:
    
    Description of the base geometry without cuts:
    {description\_template}
    
    Variation Details:
    {variation\_details}
    
    Produce exactly three labeled sections—no additional commentary:
    
    1. **User Input**  
       Formulate the user request using the base description and variation details.
    
    2. **Chain of Thought**  
       Show internal reasoning step by step:
       - Start with: ``Let me think through the requirements…"  
       - Parse counts, distances, symmetry from the Variation Details.  
       - Identify inner vs. outer face features from the base description.  
       - Plan point‐cloud density and cut placement.
    
    3. **Solution**  
       Provide a placeholder for the point‐cloud output:  
       ``Solution: {{point\_cloud}}"
}
\end{tcolorbox}
\end{tcbraster} 

\subsection{Long-term maintenance}
The APCL Lab at the University of Michigan manages the dataset and tracks user‑reported issues through the GitHub repository. For long‑term accessibility, AutoHood3D is archived on Harvard Dataverse, and our data management practices adhere to the FAIR principles \cite{wilkinson-2016_FAIR} to ensure findability, accessibility, interoperability, and reusability.


\end{document}